\documentclass[10pt,twocolumn,letterpaper]{article}
 
\usepackage[pagenumbers]{cvpr} 

\usepackage{graphicx}
\usepackage{amsmath}
\usepackage{amssymb}
\usepackage{booktabs}
\usepackage{pifont}
\usepackage{multicol}
\usepackage{multirow}

\usepackage[pagebackref,breaklinks,colorlinks]{hyperref}
\newcommand{\cmark}{\ding{51}}%
\newcommand{\xmark}{\ding{55}}%

\usepackage[capitalize]{cleveref}
\crefname{section}{Sec.}{Secs.}
\Crefname{section}{Section}{Sections}
\Crefname{table}{Table}{Tables}
\crefname{table}{Tab.}{Tabs.}

\def\cvprPaperID{859} 
\def\confName{CVPR}
\def\confYear{2023}

\begin{document}

\title{H-VFI: Hierarchical Frame Interpolation for Videos with Large Motions}

\author{Changlin Li$^{1}$\footnotemark[1] \qquad Guangyang Wu$^1$\thanks{Equal contribution} \qquad Yanan Sun$^2$ \qquad \\ Xin Tao$^{1}$ \qquad Chi-Keung Tang$^2$\thanks{Corresponding author} \qquad Yu-Wing Tai$^{1,2}$\footnotemark[2]\\
{$^1$Kuaishou Technology}  \qquad {$^2$The Hong Kong University of Science and Technology}\\
{\tt\small{changlin042@gmail.com}\hspace{0.5cm}}
{\tt\small{mulns@outlook.com}\hspace{0.5cm}}
{\tt\small{now.syn@gmail.com}\hspace{0.5cm}} \\ 
{\tt\small{jiangsutx@gmail.com}\hspace{0.5cm}}
{\tt\small{cktang@cs.ust.hk}\hspace{0.5cm}}
{\tt\small{yuwing@gmail.com}\hspace{0.5cm}}
}
\maketitle

\begin{abstract}
Capitalizing on the rapid development of neural networks, recent video frame interpolation (VFI) methods have achieved notable improvements. However, they still fall short for real-world videos containing large motions. Complex deformation and/or occlusion caused by large motions make it an extremely difficult problem in video frame interpolation. In this paper, we propose a simple yet effective solution, H-VFI, to deal with large motions in video frame interpolation. H-VFI contributes a hierarchical video interpolation transformer (HVIT) to learn a deformable kernel in a coarse-to-fine strategy in multiple scales. The learnt deformable kernel is then utilized in convolving the input frames for predicting the interpolated frame. Starting from the smallest scale, H-VFI updates the deformable kernel by a residual in succession based on former predicted kernels, intermediate interpolated results and hierarchical features from transformer. Bias and masks to refine the final outputs are then predicted by a transformer block based on interpolated results. The advantage of such a progressive approximation is that the large motion frame interpolation problem can be decomposed into several relatively simpler sub-tasks, which enables a very accurate prediction in the final results. Another noteworthy contribution of our paper consists of a large-scale high-quality dataset, YouTube200K, which contains videos depicting a great variety of scenarios captured at high resolution and high frame rate. Extensive experiments on multiple frame interpolation benchmarks validate that H-VFI outperforms existing state-of-the-art methods especially for videos with large motions.
\end{abstract}

\begin{figure}[t]
    \centering
    \includegraphics[width=0.98\linewidth]{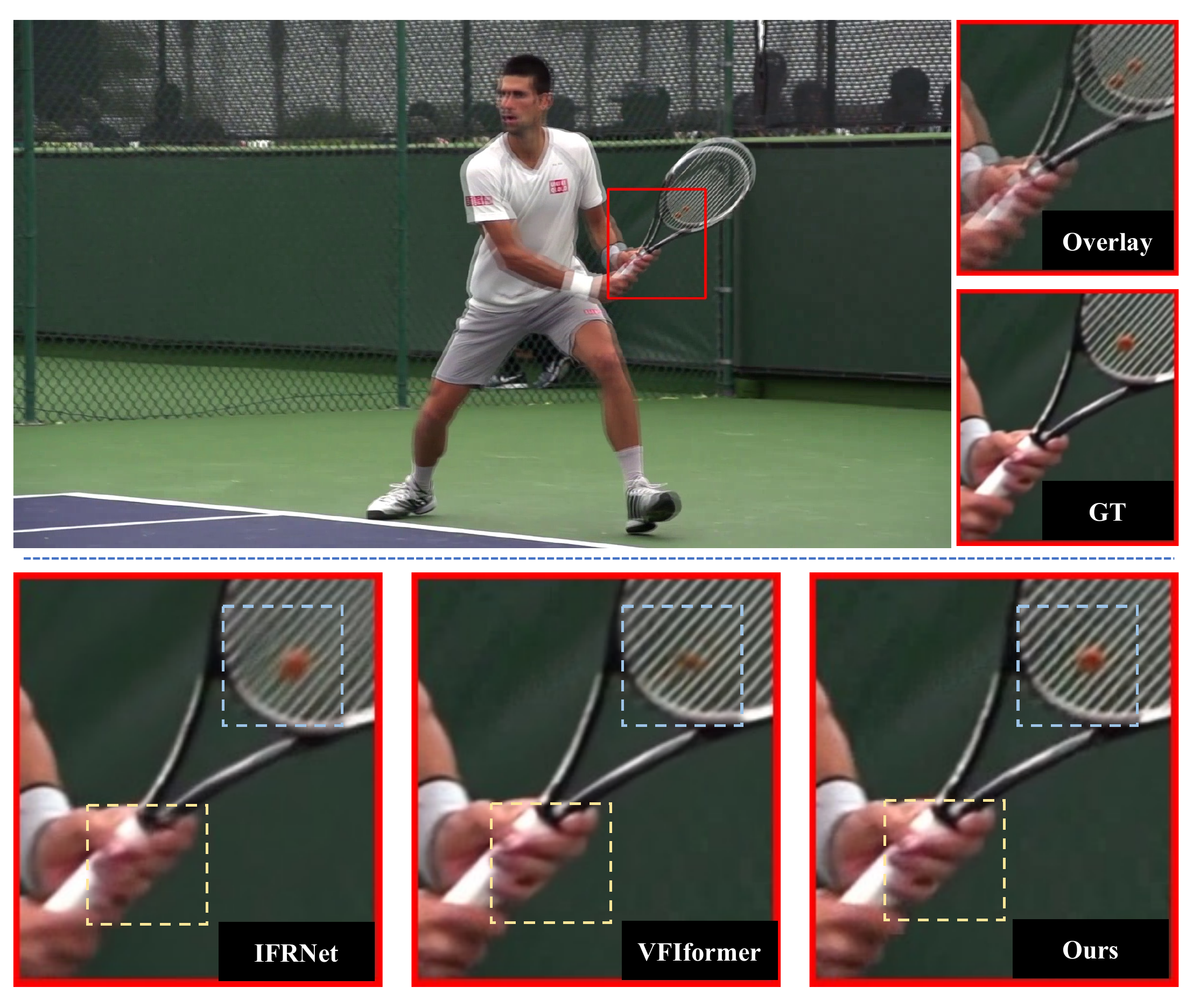}
    \vspace{-0.15in}
    \caption{Comparison on our proposed Youtube200K dataset among IFRNet~\cite{kong2022ifrnet}, VFIformer~\cite{lu2022video} and our H-VFI}\vspace{-0.25in}
    \label{fig:teaser}
\end{figure}
\vspace{-0.25in}
\section{Introduction}
\label{sec:intro}
Video Frame Interpolation (VFI) aims to synthesize intermediate frames from a low frame rate input video. It plays an important role in many applications, such as slow-motion effect simulation~\cite{superslomo2018,dain2019,contextaware2018,imnet2019,sepconv2017,voxelflow2017} and low-frame-rate video restoration~\cite{fruc:tip2018,fruc:trans1996,emmcvpr2011,edvr2019,tdan2020}. VFI is a challenging task especially for objects exhibiting large motions, not only because it is an under-constrained problem, but also because the complex deformation and occlusion can significantly hamper the performance of many existing methods that rely on accurate correspondence estimation between frames.

Existing VFI methods can be classified into optical flow-based and kernel-based methods. Optical flow-based methods~\cite{superslomo2018,dain2019,memc2021,cycle2019,contextaware2018,xvfi2021,eqvi2020,lu2022video,reda2022film, yudeep} rely on high quality explicit motion estimation, which is further integrated using convolutional networks. 
Kernel-based methods~\cite{adaconv2017,sepconv2017,imnet2019,adacof2020,pdwn,shi2022video} directly reconstruct the intermediate frames, without explicit motion estimation steps. However, both methods fall short in handling large motions in VFI. On the one hand, the performance of optical flow-based methods is limited by the accuracy of the estimated optical flows, which still contains inevitable errors, especially when dealing with non-rigid objects and occlusion. Even recent VFIformer~\cite{lu2022video}, claimed to be designed for large motion scenarios, failed in challenging cases (\emph{e.g.}~Figure~\ref{fig:teaser}). On the other hand, although kernel-based methods can achieve good performance in small motions, they cannot easily extend to handle large motions. Simply enlarging the receptive field of a network not only increases computation cost but also yields more correspondence ambiguities.

In general, we want to have a VFI method that can effectively resolve the correspondence problem in order to reconstruct an accurate intermediate frame. Inspired by the classical Lucas-Kanade algorithm~\cite{lucas-Kanade}, we propose \emph{Hierarchical Frame Interpolation method} (H-VFI), that can effectively combine the benefits of multi-scale optical flow estimation and deformable convolution in a hierarchical manner. Our key insight is that, we can use deformable kernels to estimate multiple correspondences and utilize the hierarchical update of kernel offsets to resolve  correspondence ambiguities. To ameliorate these issues, we build a hierarchical video interpolation transformer (HVIT), which produces multi-scale features by combining residual channel attention, self-attention and cross-scale attention. Specifically, we first construct a multi-frame pyramid in which the input frames are down-sampled to successive resolutions. Starting from the top layer of the pyramid, we progressively estimate the deformable kernels (DeKs) from the lowest resolution to the finest resolution. At each scale, the deformable kernel is updated with a learnt residual based on the intermediate interpolated frames and features from HVIT. In addition, we also propose a temporal gated refinement block (TGR) leveraging the interpolated frames to predict a mask and bias for final refinement. Thus, H-VFI is a simple yet effective method, which can deal with small and large motion simultaneously. Extensive experiments demonstrate that our H-VFI outperforms existing state-of-the-art methods by a large margin, especially for objects with large motion as shown in Figure~\ref{fig:teaser}. 

Furthermore, appropriate training data \eg high quality, dynamic intervals, diverse scenes etc., is indispensable in handling large-motion video interpolation. Thus, in this paper, we introduce a large-scale high quality dataset YouTube200K, which contains 200K video clips with high resolution (at least 720P) and high frame rate (30$\sim$60fps) collected from YouTube. The dataset provides a large number of videos capturing various scenes (e.g., sports, travel, street traffic) and modalities (e.g., CG animation, virtual games). Each video clip includes 30 consecutive frames with dynamic intervals setting, making the dataset flexible for different frame rate requirements. Compared with existing datasets~\cite{aim2020,vimeo,xvfi2021,cain}, our dataset is superior in scale, resolution, frame rate, motion range as well as scene diversities. 

\vspace{-0.1in}
\section{Related Work}
\label{sec:relate}
Optical flow has been a very popular approach to VFI.
Optical flow based methods address frame interpolation from the perspective of motion estimation.
They first estimate the optical flows, and then synthesize the interpolated result by adaptively blending the warped images using occlusion masks.
Specifically, SuperSlomo proposed by Jiang \etal~\cite{superslomo2018} estimated the two bi-directional optical flows and refined them using a U-Net. 
Furthermore, Reda \etal~proposed unsupervised techniques to synthesize intermediate frames using cycle consistency~\cite{reda2019}. Yuan \etal~\cite{zoom-to-check2019} warped not only input frames, but also their corresponding features. 
Liu \etal~proposed an encoder-decoder architecture to estimate 3D flow across space and time in their Deep Voxel Flow~\cite{voxelflow2017}. Moreover, CyclicGen~\cite{cycle2019} additionally used edge information~\cite{edge-detect2015} and cycle consistency loss to improve the performance of Deep Voxel Flow. 
To solve non-linear motion, Xu \etal~\cite{qvi2019} proposed QVI to exploit four consecutive frames and flow reversal filter to get the intermediate flows. Liu \etal~\cite{eqvi2020} further extended QVI with rectified quadratic flow prediction to EQVI. 
To improve the warping accuracy, Niklaus \etal~\cite{softmaxsplat2020} proposed SoftSplat to forward-warp frames and their feature map using optical flow and occlusion masks by softmax-splatting. Park \etal~\cite{bmbc} employed symmetric bilateral motion estimation to improve the accuracy of the intermediate motion. Sim \etal~\cite{xvfi2021} proposed a large motion dataset and used multi-scale optical flow estimation to capture motion. Lu \etal~\cite{lu2022video} proposed VFIformer to learn image residual and mask based on estimated flow and warped frames. Kong \etal~\cite{kong2022ifrnet} built a pyramid to reconstruct intermediate features in a coarse-to-fine manner.

Considering flow estimation as an intermediate step, other works circumvented the estimation with a single convolution process. For each output pixel, kernel-based methods learns a group of kernels in adaptive convolution to transform input frames and generate the intermediate frame. 
As a prior of kernel-based interpolation methods, AdaConv~\cite{adaconv2017} was proposed to estimate a pair of spatially-adaptive convolution kernels for each output pixel with a neural network. To reduce large memory demand, Niklaus \etal~\cite{sepconv2017} proposed SepConv that separated each 2D convolution kernel into two 1D kernels. Choi \etal~\cite{choi2020} further improved the structure of SepConv that both uni-directional and bidirectional prediction were available in video coding. 
Peleg \etal~\cite{imnet2019} modified SepConv into a multi-scale architecture and formulated interpolated motion estimation as classification by computing the center-of-mass of the convolution kernels. 
Lee \etal~\cite{adacof2020} proposed a new warping module AdaCoF by introducing a dual-frame adversarial loss to improve their performance. Shi \etal~\cite{shi2022video} introduced a transformer encoder-decoder network with four input frames to reconstruct multi-scale frames. Different from aforementioned kernel-based methods, we built a hierarchical transformer to synthesize features to supervise the deformable kernel estimation at each stage. With our hierarchical architectures and progressive training strategy, our method can simultaneously estimate multiple accurate correspondences  and therefore achieving significantly better results than previous works.


\begin{figure*}[tbp]
    \centering
    \vspace{-0.2in}
    \includegraphics[width=\linewidth]{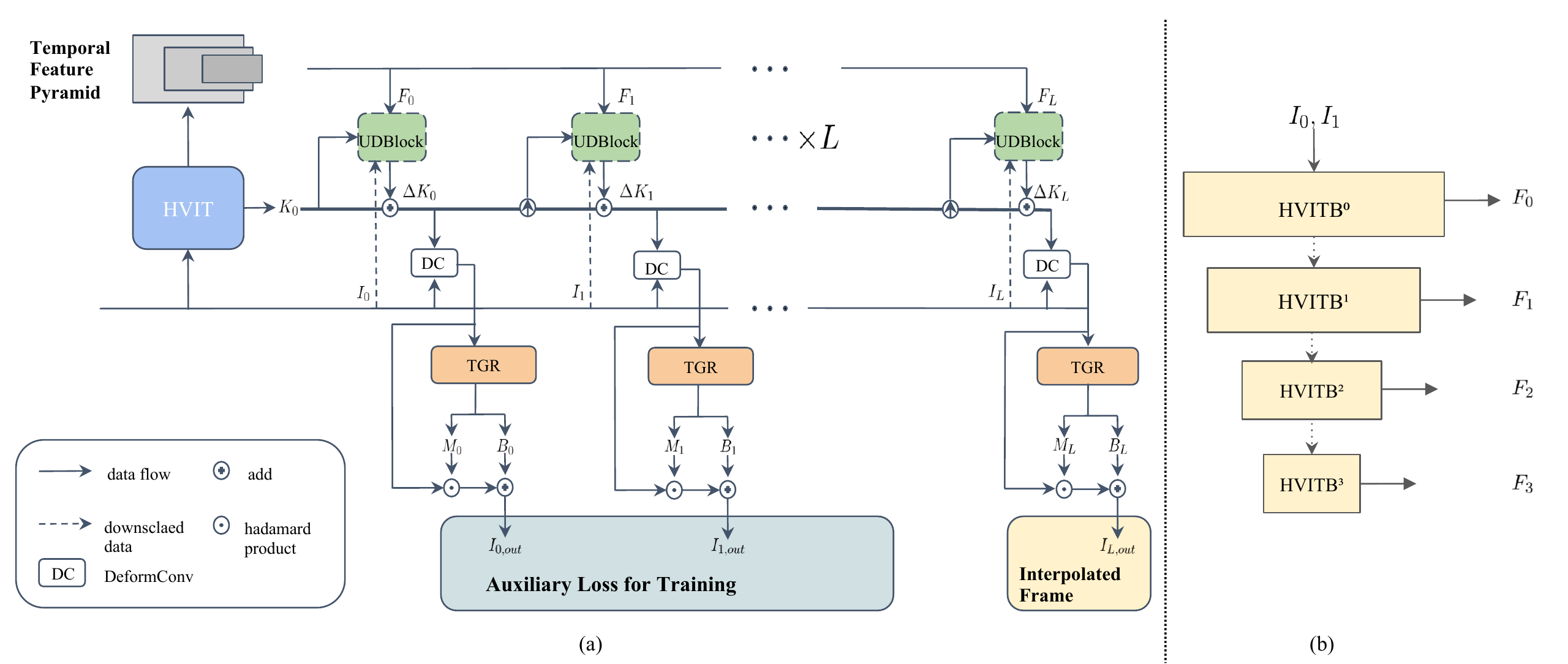}
    \vspace{-0.3in}
    \caption{(a) Framework of the proposed H-VFI. After extracting multi-scale features from HVIT, we predict deformable kernel (DeK) $\mathbf{K}$ in stages. The DeK is upscaled and corrected in succession through the predicted residual from the hierarchically connected UDBlocks, which is later illustrated in Figure~\ref{fig:udblock}. During training, we employ auxiliary loss on the intermediate interpolation results for supervising the deformable kernel learning in each scale. (b) Our HVIT combines multiple HVITB, extracting multi-scale features. Each HVITB are connected with convolution networks.}
    \label{fig:hvfi}
    \vspace{-0.2in}
\end{figure*}

\vspace{-0.1in}
\section{Deformable Convolution} 
\label{sec:dsconv}
Recent research ~\cite{dsepconv2020,shi2022video} incorporate \emph{deformable convolution} (DeformConv) ~\cite{dsepconv2020} with significant progress in frame interpolation.
Formally, DeformConv~\cite{dsepconv2020} consists of an offset $\textbf{d}$, a convolution kernel $\textbf{k}$, and a mask $\mathbf{m}$.
The resampled patch ${P}'$ is obtained by Eq.~\eqref{eq:deconv1},
\begin{align}
{P}'(x,y\,;\,\textbf{p}_i) &= {P}(x,y\,;\,\textbf{p}_i+\textbf{d}_i) \cdot {m}_i \label{eq:deconv1}
\end{align}
where $\textbf{p}_i$ denotes $i$-$th$ location in patch at $(x,y)$, $\textbf{d}_i$ and 
$m_i$ denotes the $i$-$th$ offset and the corresponding modulation scalar at $(x,y)$, $i=1,\dots,n^2$. 
The output image is obtained through convolution in Eq.~\eqref{eq:deconv2}:
\begin{align}
    \widehat{{I}}(x,y) &= {k}(x,y) * {P}'(x,y) \label{eq:deconv2}
\end{align}
where $*$ denotes the convolution operator, $k(x,y)$ denotes the convolution kernel at (x,y). The triplet,~\emph{i.e.}, the offset $\mathbf{d}$, the mask $\mathbf{m}$ and the kernel $\mathbf{k}$ together form the deformable kernel $\mathbf{K}$. 

Thus, the connection between input correlation pixels can be established by estimating $\mathbf{K}$ from a network. Whereas, directly forecast $\mathbf{K}$ simply reliance on high-resolution input frames always suffers hard to converge with blurry outputs due to large motion. Thus, in this paper, we proposed a hierarchical kernel updating strategy~\ref{update} which can build the connection between pixels with progressive updating.

\section{Method} \label{method}
Our proposed method, H-VFI, takes as input a series of $T$ input frames $\mathbf{I} = \{I^t | t=1,2,\dots,T\}$, produces intermediate frames $\widehat{\mathbf{I_s}} = \{\widehat{I_s} | s=1,2,\dots,L\}$. $L$ denotes the number of scales. Meanwhile, as mentioned above, by implementing DeformConv~\cite{dsepconv2020}, H-VFI hierarchically estimates deformable kernels $\mathbf{K} = \{K^t_s|t=1,2;s=1,2,\dots,L\}$.
Our whole framework is illustrated in Figure~\ref{fig:hvfi}(a), which includes three modules: \emph{Hierarchical video interpolation transformer (HVIT)}, \emph{Hierarchical Deformable Kernel Estimation} and \emph{Temporal Gated Refinement (TGR)}.

\subsection{Hierarchical Video Interpolation Transformer (HVIT)} \label{update}
As our hierarchical structure requires high-quality features for different scales with encoded motion information across multiple frames, inspired by \cite{liang2021swinir,chen2022activating,lu2022video}, we built a hierarchical video interpolation transformer (HVIT), shown in Figure~\ref{fig:hvfi}(b). Instead of extracting features of each scale independently as~\cite{imnet2019}, each HVIT Block (HVITB) combines residual channel attention, self-attention and cross-scale attention, such that the representation for each resolution benefits from downscaled resolution. We also introduce more multi-scale stages, so as to learn a lower resolution representation that can effectively reduce the large pixel displacements. 
\begin{figure}[t]
    \centering
    \includegraphics[width=\linewidth]{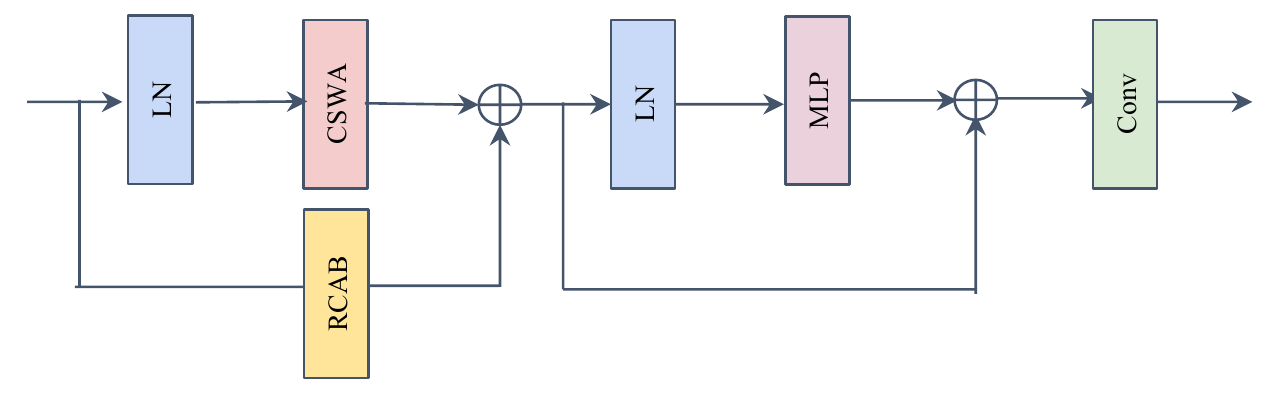}
    \vspace{-0.25in}
    \caption{Details of the structure of our HVITB. The effect of RCAB is shown in Table~\ref{table:aba_modules} and Fig~\ref{fig:ablation_residual}.}
    \label{fig:HVITB}
    \vspace{-0.2in}
\end{figure}

As shown in Figure~\ref{fig:HVITB}, we extract the temporal feature pyramid in the following: Given $T$ frames as input, we simultaneously feed them into HVIT through concatenation and extract the multi-scale spatio-temporal features $\mathbf{F} = \{{F}_s\,|\,s=1,2,\dots,L\}$. For the $s$-th HVIT Block (HVITB), the spatio-temporal feature $F_s$ is generated as 
\begin{align}
    \centering
    F_s &= \text{HVITB}^{s}({F}_{s-1}\downarrow)
\end{align}
where $\downarrow$ denotes downsample convolution. If the spatial size of input frames are $H\times W$, the size of $\widehat{I_s}$ and $F_s$ is $H/2^{L-s} \times W/2^{L-s}$. 
The temporal feature pyramid contains both high-level motion information and low-level appearance information, benefiting the network in reconstructing the intermediate frame.
More details about the HVIT can be found in the supplementary material.

\vspace{2mm}
\noindent\textbf{HVIT Block (HVITB)}\label{sec:hvitb}
Swin Transformer~\cite{liu2021Swin} has shown impressive results on low-level vision tasks. SwinIR ~\cite{liang2021swinir} built on Swin Transformer includes residual swin transformer block shows better performance. HAT~\cite{chen2022activating} introduced channel attention block based on SwinIR, which proved to get better visual representation. VFIformer~\cite{lu2022video} involved Cross-Sacle Window-based Attention (CSWA) to enlarge the receptive field.
As shown in Figure~\ref{fig:HVITB}, inspired by these works~\cite{liang2021swinir,chen2022activating,lu2022video}, we design HVITB to contain Residual Channel Attention Block (RCAB) and Cross-Scale Window-based Attention (CSWA) in parallel based on Residual Swin Transformer Block~\cite{liu2021Swin,liang2021swinir}, aiming to enlarge the receptive field for window based self-attention and thus enhance representation ability of the network. Our HVITB is computed as follows:
\begin{align}
    \mathbf{X} &= \text{CSWA}(LN(X_{in})) + \alpha(\text{RCAB}(X_{in}))
\end{align}
\vspace{-0.25in}
\begin{align}
    \mathbf{X_{out}} &= \text{MLP}(LN(\mathbf{X})) + \mathbf{X}
\end{align}
where $X_{in}$ is the input of HVITB. CSWA and RCAB respectively denote Cross-Scale Window-based Attention and Residual Channel Attention Block.
\subsection{Hierarchical Deformable Kernel Estimation}
\vspace{2mm}
\noindent\textbf{Deformable Kernel Updating.}
Given the features $\mathbf{F}$ and input frames $\mathbf{I}$, we generate and update the deformable kernels progressively at multiple scales. 
We approximate the hierarchical deformable kernel starting from an initial deformable kernel $\mathbf{K}_0$ initialized by $\mathbf{0}$. From the lowest resolution to the highest resolution, we update the deformable kernel at level $s$ as follows:  
\begin{align}
    \mathbf{K}_s &= \mathbf{K}_{s-1}\uparrow + \Delta\mathbf{K}_s. \;
    \label{eq:K_update}
\end{align}
At the $s$-th level, we initialize the deformable kernel by magnifying the offsets of $\mathbf{K}_{s-1}$ from the last level using the relative scale between two levels. As deformable kernel contains three components, each of them are updated as: 
\begin{align}
    \mathbf{O}_s^t &= \mathbf{O}_{s-1}^t\uparrow  + \Delta \mathbf{O}_s^t, \\
    \mathbf{m}_s^t &= \Delta \mathbf{m}_s^t, \\
    \mathbf{k}_s^t &= \Delta \mathbf{k}_s^t,\; \text{for}\ t=1,2
\end{align}
where $\Delta{K}_s^t=\{\Delta\mathbf{O}_s^t,\Delta\mathbf{m}_s^t,\Delta\mathbf{k}_s^t\}$. The residual offsets $\Delta\mathbf{O}_s^t$ at each stage successively update the coordinates of the corresponding pixels, while the masks and kernels are refreshed accordingly. 

\begin{figure}[t]
    \centering
    \includegraphics[width=\linewidth]{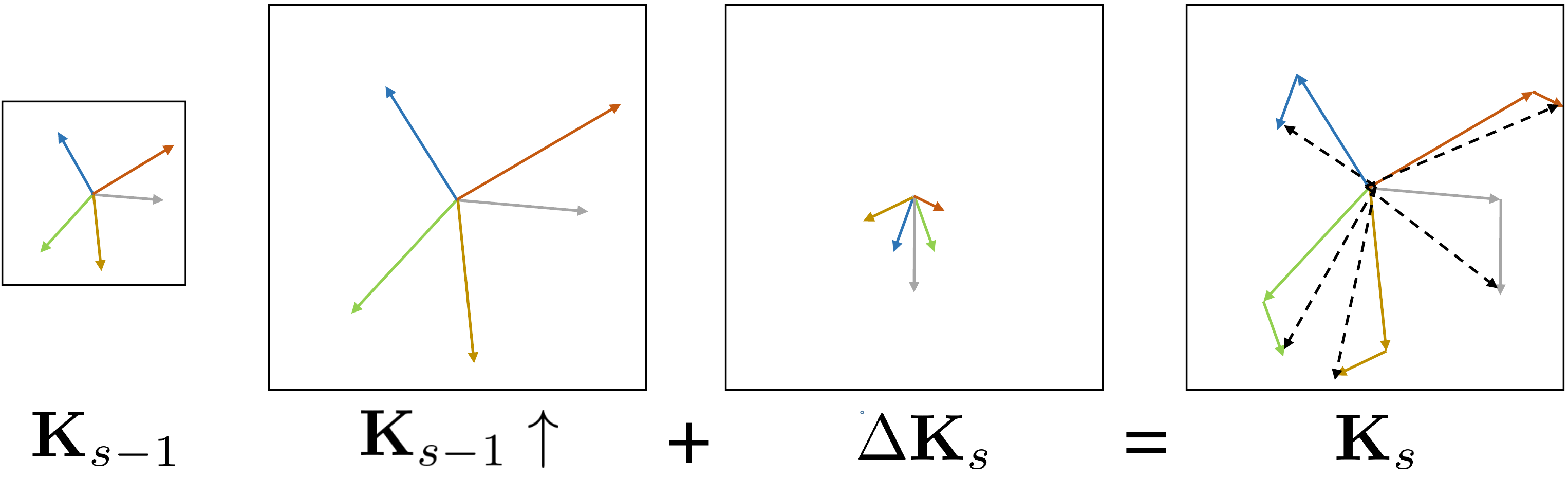}
    \vspace{-0.27in}
    \caption{Illustration of kernel updating in Eq.~\eqref{eq:K_update}. The offset of the estimated deformable kernel $\mathbf{K}_{s-1}$ from the last level is first upscaled by a factor of 2. The final deformable kernel $\mathbf{K}_s$ (marked as dash lines) is the summation of the upsampled deformable kernel $\mathbf{K}_{s-1}\uparrow$ and the learned residual $\Delta\mathbf{K}_s$.
    }
    \label{fig:kernelupdate}
    \centering
    \includegraphics[width=0.97\linewidth]{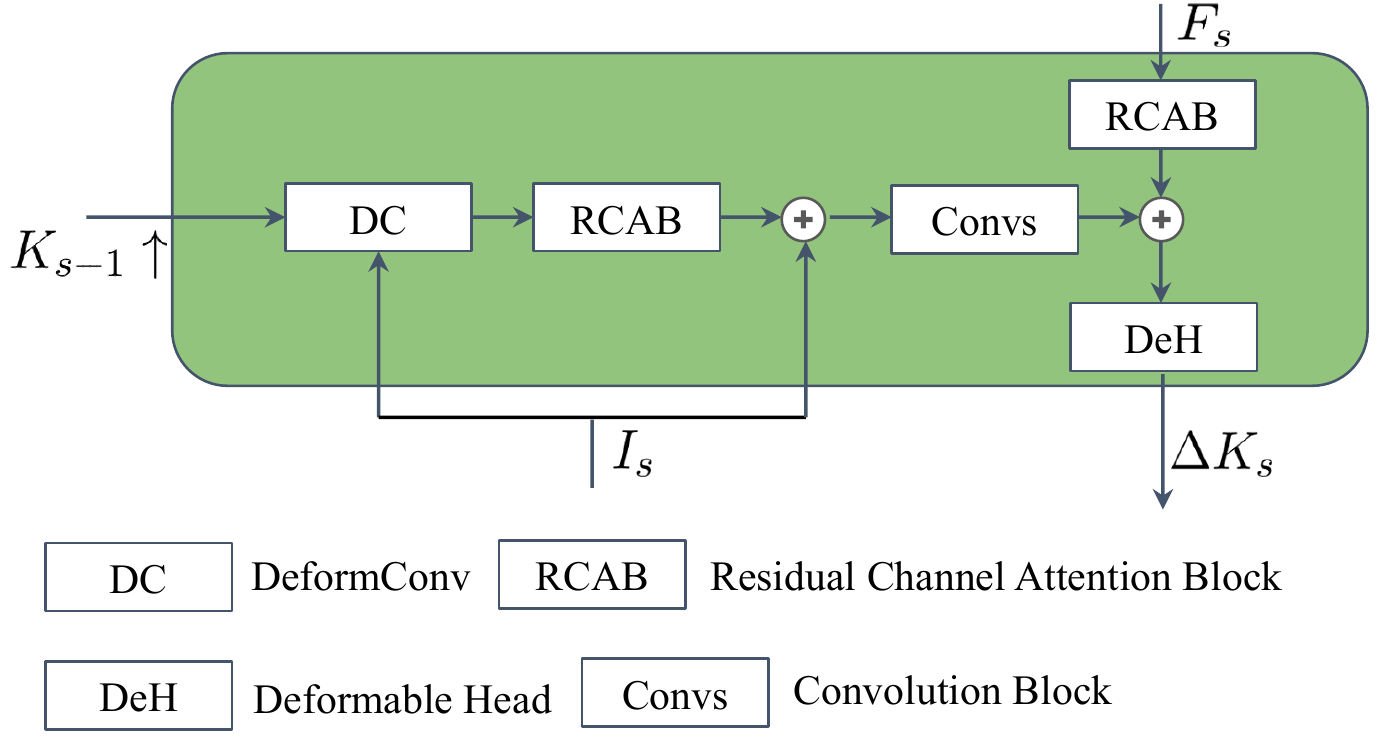}
    \vspace{-0.15in}
    \caption{The structure of s-th UDBlock. 
    }\vspace{-0.2in}
    \label{fig:udblock}
\end{figure}

The process of our progressive updating is illustrated in Figure~\ref{fig:kernelupdate}. Through the progressive updating, we approximate the deformable kernel progressively. Compared to regressing the deformable kernel individually at each level, such a decomposition strategy reduces the complexity of the large motion frame interpolation problem by narrowing the solution space in a stepwise manner.

\begin{figure*}[t] 
\begin{minipage}[b]{1.3\columnwidth}
\subfloat[]{\includegraphics[width=\linewidth]{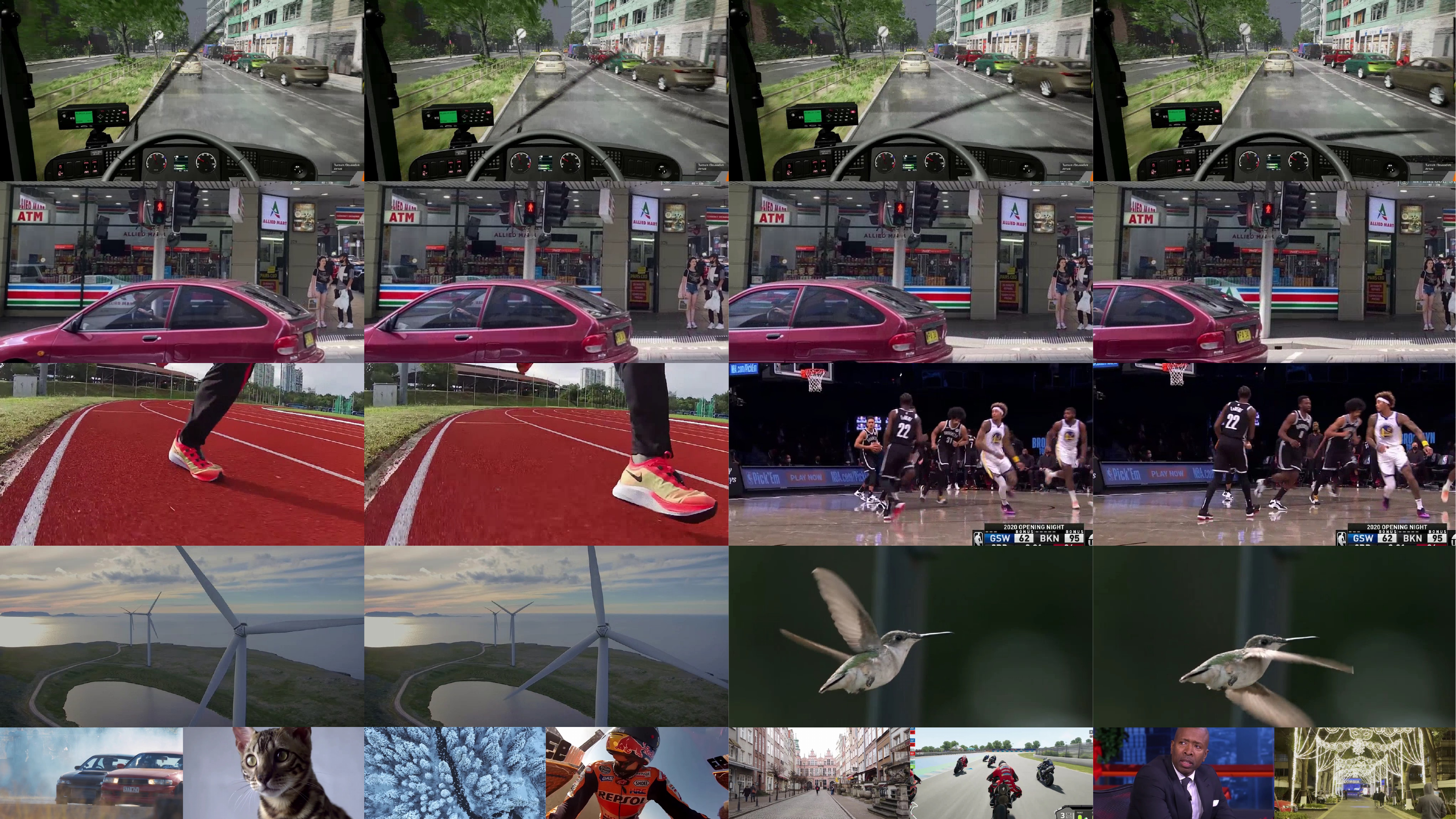}}
\end{minipage}
\hfill
\begin{minipage}[b]{0.79\columnwidth}
\subfloat[]{\includegraphics[width=1\linewidth, height=0.43\linewidth]{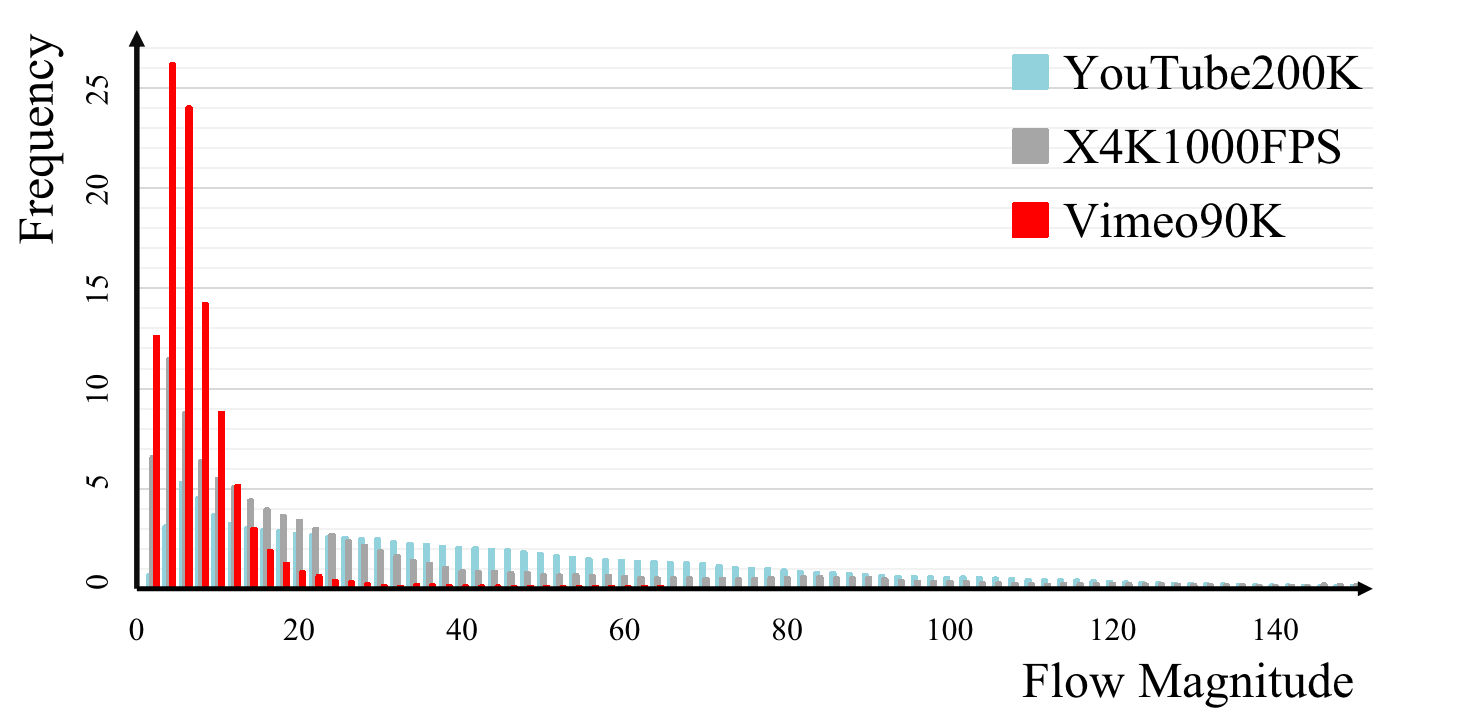}}\label{fig:histogram_1}
\hfill 

\subfloat[]{\includegraphics[width=1\linewidth, height=0.43\linewidth]{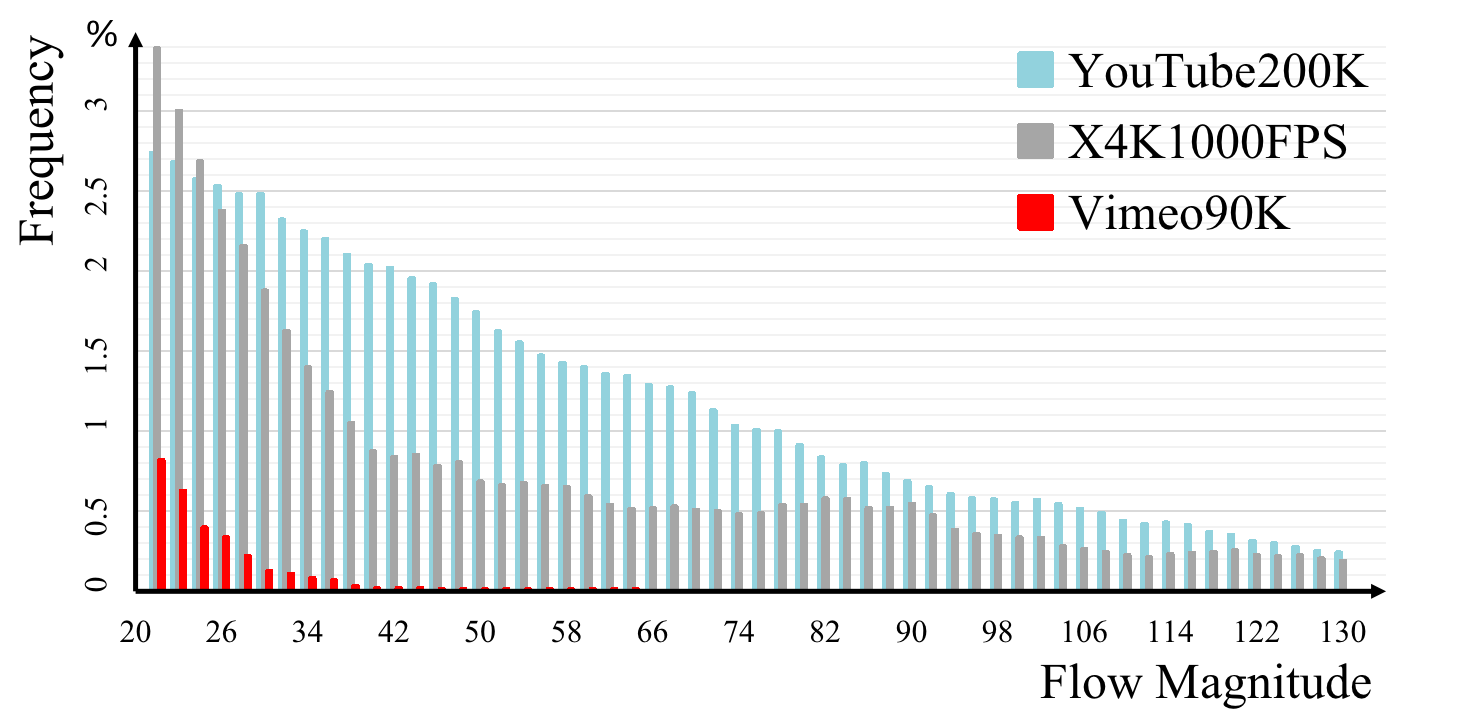}}
\hfill 
\end{minipage} 
\vspace{-0.15in}
\caption{(a) Sampled frames with large motion in various scenarios from the YouTube200K. (b) The histogram of the flow magnitudes of sampled data from YouTube200K, X4K1000FPS~\cite{xvfi2021} and Vimeo90K~\cite{vimeo}, with a zoom-in of the range of [20, 130] shown in (c).
}\vspace{-0.15in}
\label{fig:freq} 
\end{figure*}

\vspace{2mm}
\noindent\textbf{UDBlock} The deformable kernel updating block (UDBlock) is employed to estimate the residual deformable kernel $\Delta\mathbf{K}_s$. Figure~\ref{fig:udblock} illustrates the UDBlock structure. The $s$-th UDBlock takes as the input the upscaled deformable kernel $\mathbf{K}_{s-1}\uparrow$, the input frames $\mathbf{I}_s$ and the features $F_s$ at level $s$, and outputs the deformable kernel residual $\Delta \mathbf{K}_s$:
\begin{equation}
    \Delta\mathbf{K}_s = \text{UDBlock}_s\,(\mathbf{\mathbf{K}}_{s-1}\uparrow, {F}_s , \mathbf{I}_s). 
    \label{eq:delta_K}
\end{equation}
In the UDBlock, we first reconstruct the intermediate interpolated frames $\widetilde{\mathbf{I}}_{s,inter}$ for all timestamps at the $s$-th stage:
\begin{equation}
    \widetilde{\mathbf{I}}_{s,inter}= \mathit{DeformConv}(\mathbf{I}_s, \mathbf{K}_{s-1}\uparrow).\label{eq:dsc}
\end{equation}
Next, $\widetilde{\mathbf{I}}_{s,inter}$ and $\mathbf{I}_s$ are concatenated and then fed into $M$ channel attention blocks (CABs) for extracting its feature embeddings. Afterward, we concatenate the extracted features with the spatio-temporal features ${F}_s$ obtained from the HVIT. Finally, the concatenated features are fed into a deformable head (DeH) to estimate the residual of deformable kernels. 
The deformable head consists of multiple prediction heads in parallel, producing offset, kernel and mask prediction respectively for each frame as in~\cite{dsepconv2020}. We use the interpolated image $\widetilde{\mathbf{I}}_{s,inter}$ instead of $\widetilde{\mathbf{I}}_{s-1,inter}\uparrow$ to estimate $\Delta\mathbf{K}_s$ because $\widetilde{\mathbf{I}}_{s,inter}$ contain high resolution details which is essential for precise residual estimation. During training, we produce an interpolated output $\widehat{I}_{s,out}$ at each stage. Specifically, after obtaining $\mathbf{K}_s$ by Eq.~\eqref{eq:K_update}, we reconstruct the intermediate interpolated frame $\widetilde{I}_{s}$ as follows for performing auxiliary supervision,
\begin{equation}
    \widetilde{I}_{s}^{t} = \mathit{DeformConv}(I_s^t, {K}_{s}^t)
\end{equation}

\vspace{2mm}
\noindent\textbf{Temporal Gated Refinement (TGR)}
As direct fusion of the output of deformable convolution $\widetilde{I}_{s}$ by averaging causes blurry or areas dislocation in the output, inspired by~\cite{kong2022ifrnet,lu2022video}, we proposal Temporal Gated Refinement (TGR), which invokes $\widetilde{I}_{s}$ to predict a soft mask $M_s$ and bias $\Delta I_s$. Thus, the final output can be reconstructed as follows:
\begin{equation}
    \widehat{I}_{s,out} = \widetilde{I}_{s}^{0} \odot M_s + \widetilde{I}_{s}^{1} \odot (1 - M_s) + \Delta I_s
\end{equation}
We use an HVIT block as the architecture of TGR.

\vspace{2mm}
\noindent\textbf{Loss Function.}
We adopt the multi-scale reconstruction loss function and Cencus loss\cite{meister2018unflow} with $7\times7$ patches to train the framework. 
Note that the $\widehat{I}_{L,out}$ is the final output interpolated frame. At each stage, we compute the loss between $\widehat{I}_{s,out}$ and the corresponding downsampled groundtruth $I_{GT}$. The total loss $\mathcal{L}$ sums them up to:
\begin{align}
\!\!\!\!\mathcal{L} &\!=\! \sum_{s=1}^{L} (\|\, \widehat{I}_{s,out}-I_{GT}\downarrow_s \|_1\!+\!\mathcal{L}_{cen}(\widehat{I}_{s,out},I_{GT}\downarrow_s))
\end{align}
where $\widehat{I}_{s,out}$ is the generated frame after TGR. The $I_{GT}\downarrow_s$ denotes the downsampled ground truth frame with the same spatial size as $\widehat{I}_{s,out}$. 

\section{YouTube200K Dataset}
We construct a dataset called YouTube200K, consisting 200K video clips for the research on large motion video frame interpolation and other related tasks.

\vspace{2mm}
\noindent\textbf{Dataset Collection.} The widely used Vimeo90K dataset contains 4278 videos and 73171 triplets are sampled for VFI tasks. However, this dataset can no longer meet the high resolution and high frame rate requirements of many contemporary videos. To acquire high quality videos for video processing, previous methods~\cite{aim2020,xvfi2021} collect videos on their own 
in limited scale and content. 
Alternatively, we collect high-quality and diverse videos from YouTube, focusing on videos in high resolution (720P at least) and high frame rate (mostly 60FPS). The dataset can be divided into two subsets. Youtube200k-slowmo encompasses slow-motion videos, including extreme sports, motorcycle races, tennis and parkour etc. Youtue200k-normal includes various scenes and modalities, including live shows, CG animation, game interface, etc, with their scene contents covering road traffic, travel logs, and natural scenery, etc. 
\begin{table*}[t]
    \centering
    \resizebox{\textwidth}{!}{%
    \begin{tabular}{c|c|c|c|c|c|c}
    \hline\hline
    \multicolumn{1}{c|}{\multirow{2}{*}{Method}} &
     \multicolumn{1}{c|}{\multirow{1}{*}{Vimeo90K}} &
     \multicolumn{1}{c|}{\multirow{1}{*}{SNU-FILM (Hard)}} &
     \multicolumn{1}{c|}{\multirow{1}{*}{SNU-FILM (Extreme)}} &
     \multicolumn{1}{c|}{\multirow{1}{*}{YouTube200K (I3)}} &
     \multicolumn{1}{c|}{\multirow{1}{*}{YouTube200K (I4)}} &
     \multicolumn{1}{c}{\multirow{1}{*}{Params}}\\
     \multicolumn{1}{c|}{\multirow{1}{*}{}} &
     \multicolumn{1}{c|}{\multirow{1}{*}{PSNR/SSIM}} & 
     \multicolumn{1}{c|}{\multirow{1}{*}{PSNR/SSIM}} & 
     \multicolumn{1}{c|}{\multirow{1}{*}{PSNR/SSIM}} & 
     \multicolumn{1}{c|}{\multirow{1}{*}{PSNR/SSIM}} & 
     \multicolumn{1}{c|}{\multirow{1}{*}{PSNR/SSIM}} &
     \multicolumn{1}{c}{\multirow{1}{*}{(M)}} \\ \hline
    EQVI~\cite{eqvi2020} &
      34.05 / 0.967 &
      27.65 / 0.913 &
      24.55 / 0.853 &
      26.08 / 0.893 &
      25.56 / 0.886 &
      42.8\\
    CAIN~\cite{cain} &
      34.65 / 0.973 &
      29.86 / 0.929 &
      24.69 / 0.851 &
      28.55 / 0.913 &
      27.14 / 0.898 &
      42.78\\
    AdaCoF~\cite{adacof2020} &
      34.47 / 0.973 &
      29.46 / 0.924 &
      24.31 / 0.844 &
      27.90 / 0.904 &
      26.48 / 0.886 &
      21.8\\
    EDSC~\cite{edsc2020} &
      34.84 / 0.975 &
      29.60 / 0.926 &
      24.39 / 0.843 &
      28.23 / 0.909 &
      26.78 / 0.892 &
      8.95\\
    CDFI~\cite{cdfi2021} &
      34.89 / 0.969 &
      29.76 / 0.928 &
      24.55 / 0.848 &
      28.32 / 0.910 &
      26.87 / 0.893 &
      5.0\\
    FLAVR~\cite{flavr} &
      36.30 / 0.975 &
      30.87 / 0.942 &
      25.90 / 0.876 &
      28.84 / 0.916 &
      27.44 / 0.901 &
      42.06\\
    XVFI~\cite{xvfi2021} &
      35.21 / 0.970 &
      29.43 / 0.928 &
      24.02 / 0.841 &
      27.99 / 0.911 &
      26.83 / 0.899 &
      5.61\\    
    ABME~\cite{abme2021} &
      36.18 / 0.980 &
      30.58 / 0.936 &
      25.42 / 0.864 &
      28.22 / 0.912 &
      26.80 / 0.897 &
      18.1\\      
    Softsplat~\cite{softmaxsplat2020} &
      35.76 / 0.972 &
      - &
      - &
      - &
      - &
      -\\    
    IFRNet-L~\cite{kong2022ifrnet} &
      36.20 / 0.981 &
      30.63 / 0.937 &
      25.27 / 0.861 &
      \textcolor{blue}{29.07 / 0.923} &
      27.68 / 0.910 & 
      19.7\\
    VFIformer~\cite{lu2022video} &
      \textcolor{red}{36.50 / 0.982} &
      30.67 / 0.938 &
      25.43 / 0.864 &
      29.01 / 0.922 &
      27.57 / 0.909 &
      24.1 \\ 
    DBVI~\cite{yudeep} &
      36.17 / 0.977 &
      \textcolor{red}{31.68 / 0.953} &
      \textcolor{red}{25.90 / 0.876} &
      - &
      - &
      21.69 \\ \hline    
    H-VFI (Ours) &
      {36.08 / 0.980} &
      {30.11 / 0.934} &
      {24.88 / 0.859} &
      {28.71 / 0.919} &
      {27.33 / 0.906} &
      13.0\\ 
    H-VFI-Large (Ours) &
      \textcolor{blue}{36.37 / 0.981} &
      {31.31 / 0.950} &
      {25.59 / 0.873} &
      {29.06 / 0.923} &
      \textcolor{blue}{27.71 / 0.910} &
      22.7\\ 
    H-VFI-Large$\ast$ (Ours) &
      {36.35 / 0.981} &
      \textcolor{blue}{31.40 / 0.950} &
      \textcolor{blue}{25.71 / 0.874} &
      \textcolor{red}{29.69 / 0.931} &
      \textcolor{red}{28.49 / 0.923} &
      22.7\\ \hline \hline
    \end{tabular}%
    }
    \vspace{-0.1in}
    \caption{Quantitative comparisons on five datasets. The numbers in \textcolor{red}{red} and \textcolor{blue}{blue} represent the best and the second best performance, respectively.
    We include the Hard and Extreme mode of the SNU-FILM dataset, and the YouTube200K where the interval is set to 3 and 4 (abbreviated by I3, I4) for large motion comparisons, while the Vimeo dataset contains smaller motion magnitude. $\ast$ denotes the model trained on Youtube200K.}
    \label{tab:main}
    \vspace{-0.22in}
\end{table*}

To convert the long videos to clips appropriate for training and testing, we adopt the following procedure. First, we roughly group all the collected videos into several categories according to their video contents. Then we select videos in balanced number from different categories. The selected videos are randomly cut into 30-frame shots. Yet, some of these clips are unqualified due to the appearance of jump cuts. Therefore, we use a simple threshold analysis to filter out these defective clips containing jump cuts, in order to maintain the continuity of the video clip contents. 
Finally, a total of 7,824 videos  meet our high requirements. 
Figure~\ref{fig:freq}(a) visualizes some samples with large motion and variety of scenarios. We partition all the clips into two subsets, 6824 and 1,000 clips served as the training and testing split respectively.

\vspace{2mm}
\noindent\textbf{Comparison with Existing Datasets}
Different from other datasets, Youtube200K provides two subsets: Youtube200K-slowmo and Youtube200K-normal in multi-resolution video form with dynamic interval rather than fixed-size images with invariant interval. Leveraged the advantages of that, Youtube200K can accommodates diversified training requirements. Table~\ref{tab:dataset} compares our YouTube200K with existing and widely used datasets, e.g., Vimeo90K~\cite{vimeo}, and X4K1000FPS~\cite{xvfi2021} in different aspects, revealed that YouTube200K is superior to X4K1000FPS in quantity and to Vimeo90K in resolution and frame rate. In addition, we analyze the flow magnitudes of the three datasets, computed between the first frame and the last frame in a given video using RAFT~\cite{raft}. To be more specific, we sample the same number (180M) of motion vectors for the three datasets for fairness of comparison. Then we plot the histogram of flow magnitudes for the sampled motion vectors as shown in Figure~\ref{fig:freq}(b)--(c). Compared to Vimeo90K and X4K1000FPS, our YouTube200K contains very rich motion information and thus better support research on large motion video frame interpolation.

\section{Experiments}
\subsection{Implementation Details}
\noindent\textbf{Training Schedule.}
In HVIT, the number of stages $L$ is set to 4. 
The size of deformable kernel $n$ is set to 5, indicating that the deformable kernel  finds 25 corresponding pixels for each output pixel. Inside the UDBlock, we stack $M=4$ channel attention blocks to extract local features.  
We use the AdamW optimizer~\cite{adam} with a learning rate of $\eta = 3 \times 10^{-4}$ and cosine learning rate decay. We set the mini-batch size 64 and train the network for 600 epochs with 8 NVIDIA V100 GPUs. 

\begin{table}[t]
    \centering
    \begin{tabular}{l|c|c|c}
    \hline\hline
        Dataset & Quantity & Resolution & FPS \\
        \hline
        X4K1000FPS~\cite{xvfi2021} & 4,408 & 4096 $\times$ 2160 & 1,000\\
        Vimeo90K~\cite{vimeo} & 73,171 & 448 $\times$ 256 & $\leq$30 \\
        YouTube200K & 246,600 & 1280 $\times$ 720$\ \ $ & 60 \\
    \hline\hline
    \end{tabular}
    \vspace{-0.1in}
    \caption{Comparisons among widely used training datasets for frame interpolation and our YouTube200K.}\vspace{-0.12in}
    \label{tab:dataset}
\end{table}

\begin{figure*}[t]
    \centering
    \includegraphics[width=\linewidth]{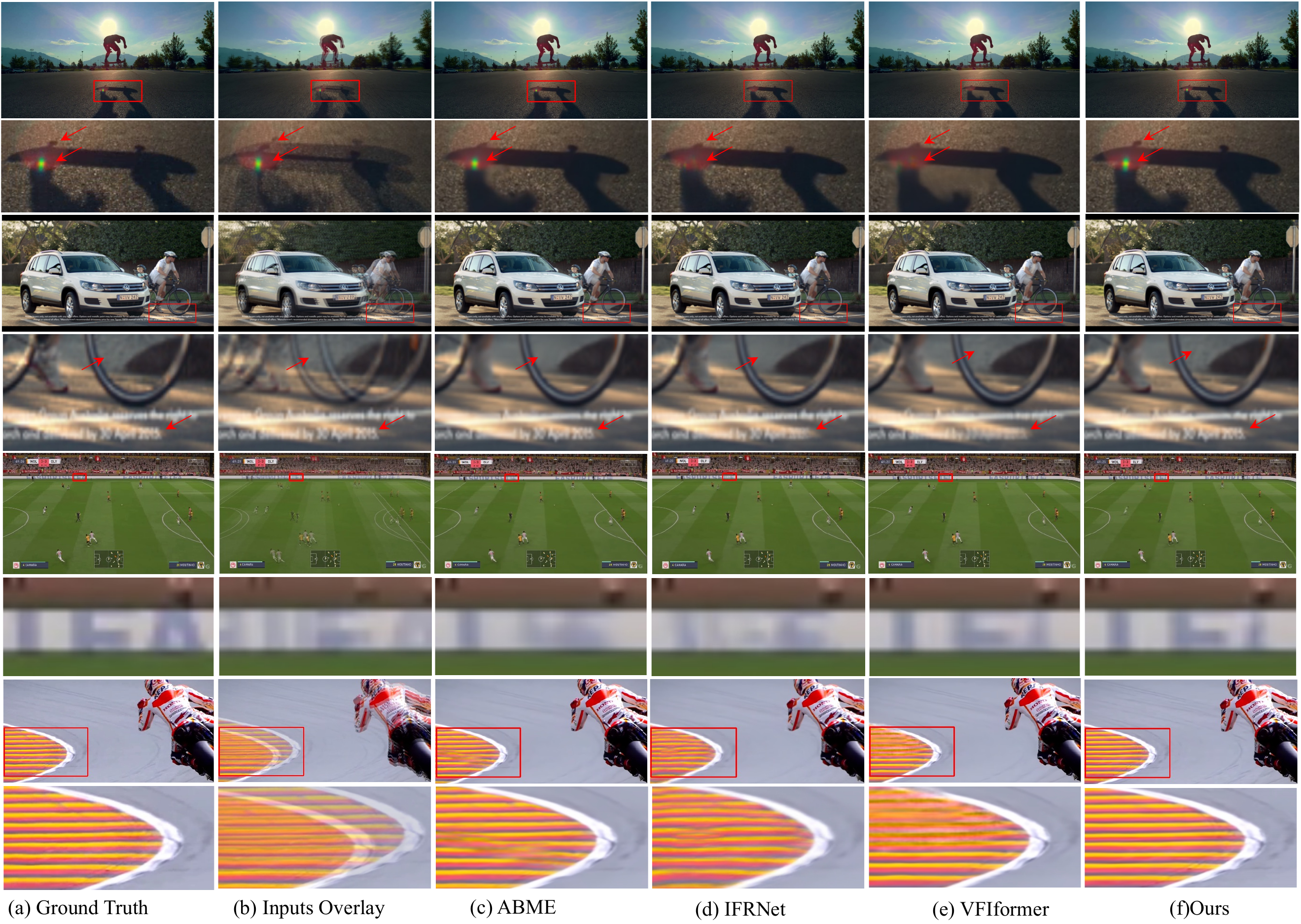}
    \vspace{-0.22in}
    \caption{Qualitative comparisons of our method H-VFI with ABME~\cite{abme2021}, IFRNet~\cite{kong2022ifrnet} and VFIformer~\cite{lu2022video}}
    \label{fig:visual1} \vspace{-0.2in}
\end{figure*}

\begin{table}[htbp]
    \centering
    \resizebox{\linewidth}{!}{%
    \begin{tabular}{c|c|c|c|c}
    \hline\hline
    \multicolumn{1}{c|}{\multirow{2}{*}{Method}} & 
    \multicolumn{1}{c|}{\multirow{1}{*}{Y.T.-I1}} &
    \multicolumn{1}{c|}{\multirow{1}{*}{Y.T.-I2}} &
    \multicolumn{1}{c|}{\multirow{1}{*}{Y.T.-I3}} &
    \multicolumn{1}{c}{\multirow{1}{*}{Y.T.-I4}} \\
    \multicolumn{1}{c|}{\multirow{1}{*}{ }} & 
    \multicolumn{1}{c|}{\multirow{1}{*}{PSNR/SSIM}} &
    \multicolumn{1}{c|}{\multirow{1}{*}{PSNR/SSIM}} &
    \multicolumn{1}{c|}{\multirow{1}{*}{PSNR/SSIM}} &
    \multicolumn{1}{c}{\multirow{1}{*}{PSNR/SSIM}} \\
    \hline
    ABME~\cite{abme2021} & 34.18 / 0.959 &	{31.14 / 0.942} &{28.22 / 0.913} & {26.80 / 0.897} \\
    IFRNet~\cite{kong2022ifrnet} & \textcolor{blue}{34.71 / 0.961} & \textcolor{blue}{32.02 / 0.950} & 29.07 / 0.923 & 27.68 / 0.910 \\
    VFIformer~\cite{lu2022video} & {34.61 / 0.961} & 31.90 / 0.949 & 29.01 / 0.922 & 27.57 / 0.909 \\
    \hline
    H-VFI (Ours) & {34.45 / 0.960} &{31.62 / 0.946} &{28.71 / 0.919} &{27.33 / 0.906} \\
    H-VFI-Large (Ours) & {34.70 / 0.961} &\textcolor{blue}{32.02 / 0.950} &	\textcolor{blue}{29.06 / 0.923} & \textcolor{blue}{27.71 / 0.910} \\
    H-VFI-Large$\ast$(Ours) & \textcolor{red}{34.83 / 0.963} &	\textcolor{red}{32.21 / 0.952} &	\textcolor{red}{29.69 / 0.931} &\textcolor{red}{28.49 / 0.923} \\
    \hline\hline
    \end{tabular}%
    }
    \vspace{-0.1in}
    \caption{Comparisons on YouTube200K with different intervals $i=1,2,3,4$ (abbreviated with Y.T.-I1, -I2, -I3, -I4). Larger interval represents larger motion and lower input frame rate. $\ast$ denotes the model trained on Youtube200K.}
    \label{tab:YT}
    \vspace{-0.2in}
\end{table}

\vspace{2mm}
\noindent\textbf{Datasets.}
We use Vimeo90K~\cite{vimeo} training set for training HVIT/HVIT-Large from scratch. Meanwhile, we also train HVIT-Large on Youtube200K for comparison. To efficiently load the video frames without pre-decoding videos into images, we adopt the NVIDIA DALI~\cite{dali} to process videos on GPUs. Data augmentation is carried out in the temporal and spatial dimensions. For temporal dimension, for each sample with 30 frames, we select the 14-$th$ frame as the label and the indices of input frames are $14-i$, $14$, $14+i$ where the interval $i \in [2,6]$ is selected in normal video / slow motion video. Different intervals represent different input frame rates. We also randomly reverse the chronological order. For spatial dimension, the video is randomly cropped into patches of size $128 \times 128$ and flipped horizontally or vertically.

For evaluation we test on 
Vimeo90K~\cite{vimeo}, SNU-FILM~\cite{cain}, and our YouTube200K. For Vimeo90K, we use the triplet testing set in which 3782 video sequences have 3 frames each. For SNU-FILM, we test the model on the easy, medium, hard and extreme sets. The YouTube200K has 1000 video clips for testing, each of which contains 30 frames. We select the 14-$\mathit{th}$ frame as the label, and the ($14-i$, $14+i$) -$\mathit{th}$ frames as the inputs. The interval $i$ is set to~4 for large motion. We  compare the results with different intervals ($i=1,2,3,4$) on YouTube200K. 

\subsection{Evaluation}
We compare H-VFI with conventional algorithms: EQVI~\cite{eqvi2020},  CAIN~\cite{cain}, AdaCoF~\cite{adacof2020}, EDSC~\cite{edsc2020}, CDFI~\cite{cdfi2021}, FLAVR~\cite{flavr}, XVFI~\cite{xvfi2021}, ABME~\cite{abme2021}, Softsplat~\cite{softmaxsplat2020}, IFRNet-L~\cite{kong2022ifrnet}, VFIformer~\cite{lu2022video} and DBVI~\cite{yudeep}. 
Table~\ref{tab:main} compares the average PSNR / SSIM scores. The SNU-FILM (Hard, Extreme) and YouTube200K (I3, I4) satisfy the large motion criterion, where our H-VFI outperforms the other methods. In Vimeo90K which contains smaller motion magnitude, our method still produces comparable results.

To better show the scalability of our method on both large and small motion datasets, we also compare these methods on the same dataset with different motion magnitude. In Table~\ref{tab:main} and Table~\ref{tab:YT}, we set different time intervals between consecutive frames where $i=1,2,3,4$, respectively. For each interval, our H-VFI shows outstanding performance against other methods, which indicates that H-VFI can faithfully interpolate challenging videos.

\begin{figure}[t]
    \centering
    \includegraphics[width=0.98\linewidth]{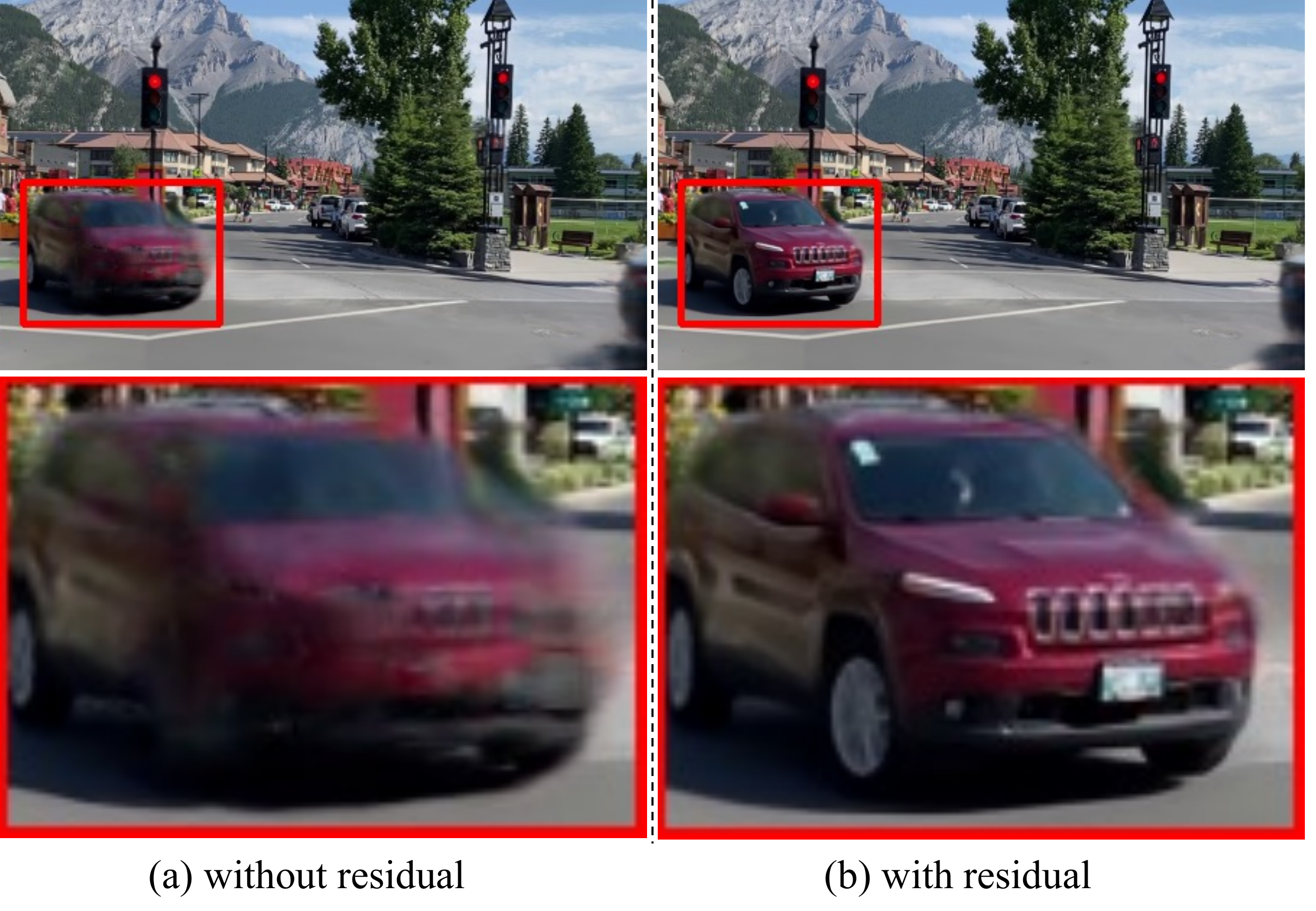}
    \vspace{-0.15in}
    \caption{Comparisons between the two deformable kernel estimaton strategies, i.e., directly regressing the deformable kernel (denoted as ``without residual"), and  estimating the deformable kernel residual (denoted as ``with residual").}
    \label{fig:ablation_residual}
    \vspace{-0.2in}
\end{figure}

\subsection{Analysis}
\noindent\textbf{Comparisons with Existing Methods.}
Figure~\ref{fig:visual1} shows four sets of result samples to compare the visual quality among flow-based methods~\cite{kong2022ifrnet,lu2022video} and kernel-based method~\cite{abme2021}. Compared with flow-based methods, H-VFI clearly recovers textures more correctly. Meanwhile, H-VFI recovers more details than kernel-based ABME~\cite{abme2021}. 
Even though VFIformer~\cite{lu2022video} achieves a higher PSNR value, it fails on reconstructing correct image content with flow estimation. Our better design benefits explicit auxiliary supervision and produces the sharpest interpolated result. Compared to the optical flow, the deformable kernel is much more robust to complex textures and occlusions, which explains ABME and H-VFI generate less artifacts.

\begin{figure}[t]
    \centering
    \includegraphics[width=\linewidth]{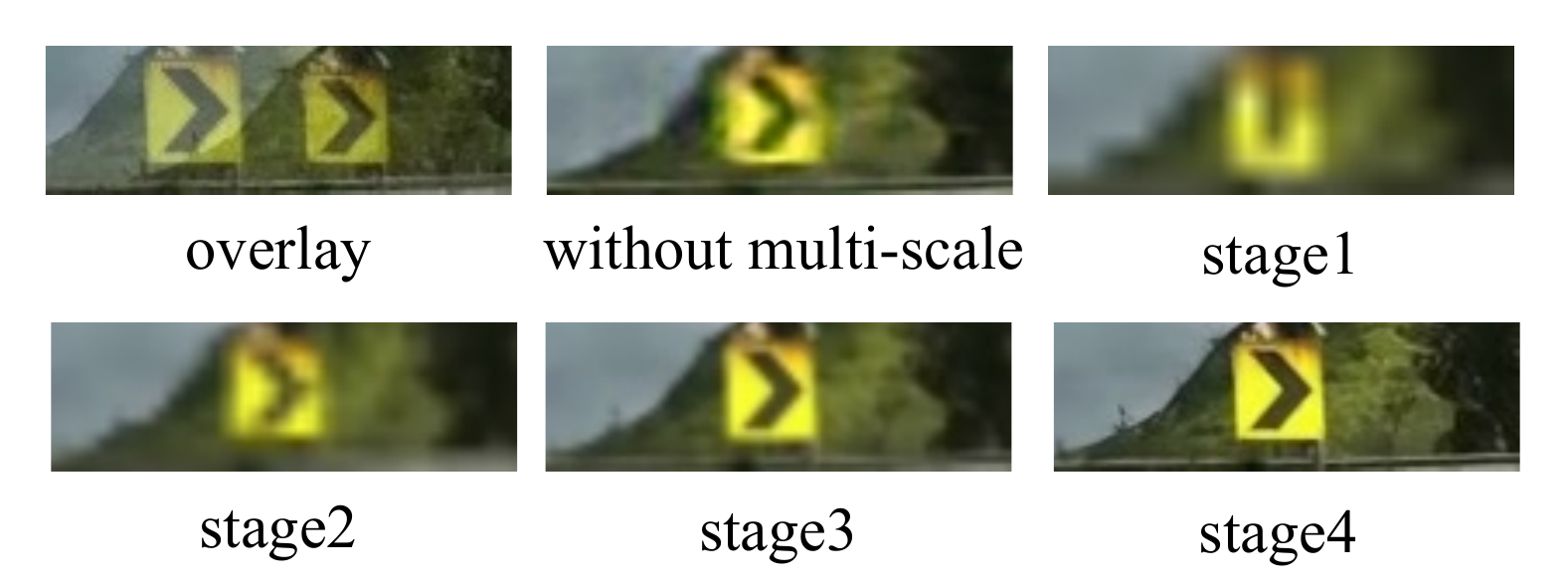}
    \vspace{-0.3in}
    \caption{Output of our baseline without multi-scale scheme and outputs from different stages of H-VFI.}
    \label{fig:ablation_visual_scales}
    \vspace{-0.2in}
\end{figure}
\vspace{2mm}
\noindent\textbf{Effect of learning the residual.}\label{sec:residual} In the hierarchical deformable kernel learning, we have two strategies to estimate the deformable kernel $\mathbf{K}_s$ in each scale: 1) directly regress the deformable kernel based on $\mathbf{K}_{s-1}$ and $I_s$, 2) estimate the  residual $\Delta\mathbf{K}_s$ and to add to $\mathbf{K}_{s-1}$. Compared to the first strategy, learning a residual benefits from not only the multi-scale information but also the complexity reduction. That is, in each scale, the UDBlock finds the correspondence within the surrounding region at the end of $\mathbf{K}_{s-1}$, instead of the whole surrounding region covering from the start to the end of $\mathbf{K}_{s-1}$. This effectively narrows the solution space by a large margin, and thus greatly reduces the computational complexity. We show the comparisons between the two strategies in Figure~\ref{fig:ablation_residual} and Table ~\ref{table:aba_modules}. As shown, it is much more difficult to learn the correspondence of large motion via directly regressing the deformable kernel, while  residual learning produces a more precise estimation.

\vspace{2mm}
\noindent\textbf{Effect of multi-scale structure. }\label{sec:multiscale}
Here, we visualize the intermediate results to show the importance of multi-scale structure for capturing large motions. Figure~\ref{fig:ablation_visual_scales} shows that it is very challenging to interpolation approaches that directly learn large motion without a multi-scale scheme, where the wrong offset leads to unsightly visual results. On the contrary, in the ``stage 1" result of Figure~\ref{fig:ablation_visual_scales}, down-sampling the image can effectively alleviate this problem. Even though the image details are lost at this low resolution, the network captures the correct underlying motion. Using the coarse-to-fine structure, we can progressively improve the image details and generate high-quality intermediate frames.
\begin{table}[t]
    \setlength{\belowcaptionskip}{0pt}
    \centering
    \scalebox{1}{
        \begin{tabular}{l|cc|c}
            \hline 
            \multicolumn{1}{c|}{} & \multicolumn{1}{c}{RCAB} & \multicolumn{1}{c}{residual updating} & \multicolumn{1}{c}{Vimeo90K} \\
            \hline 
            Model 1 & \xmark & \xmark & 36.05/0.980 \\ 
            Model 2 & \xmark & \cmark & 36.14/0.980  \\ 
            Model 3 & \cmark & \cmark & 36.35/0.981  \\ 
            \hline 
    \end{tabular}}
    \vspace{-0.1in}
    \caption{Ablation study on RCAB and learning the residual. }
    \label{table:aba_modules}
    \vspace{-0.19in}
\end{table}

\vspace{2mm}
\noindent\textbf{Effect of Residual Channel Attention Block}
In section~\ref{sec:hvitb}, different from other transformer based models, we add Residual Channel Attention Block (RCAB) in our HVITB to enhance visual quality. Table~\ref{table:aba_modules} shows the effect of RCAB in HVIT.  

\begin{figure}[t]
    \centering
    \includegraphics[width=\linewidth]{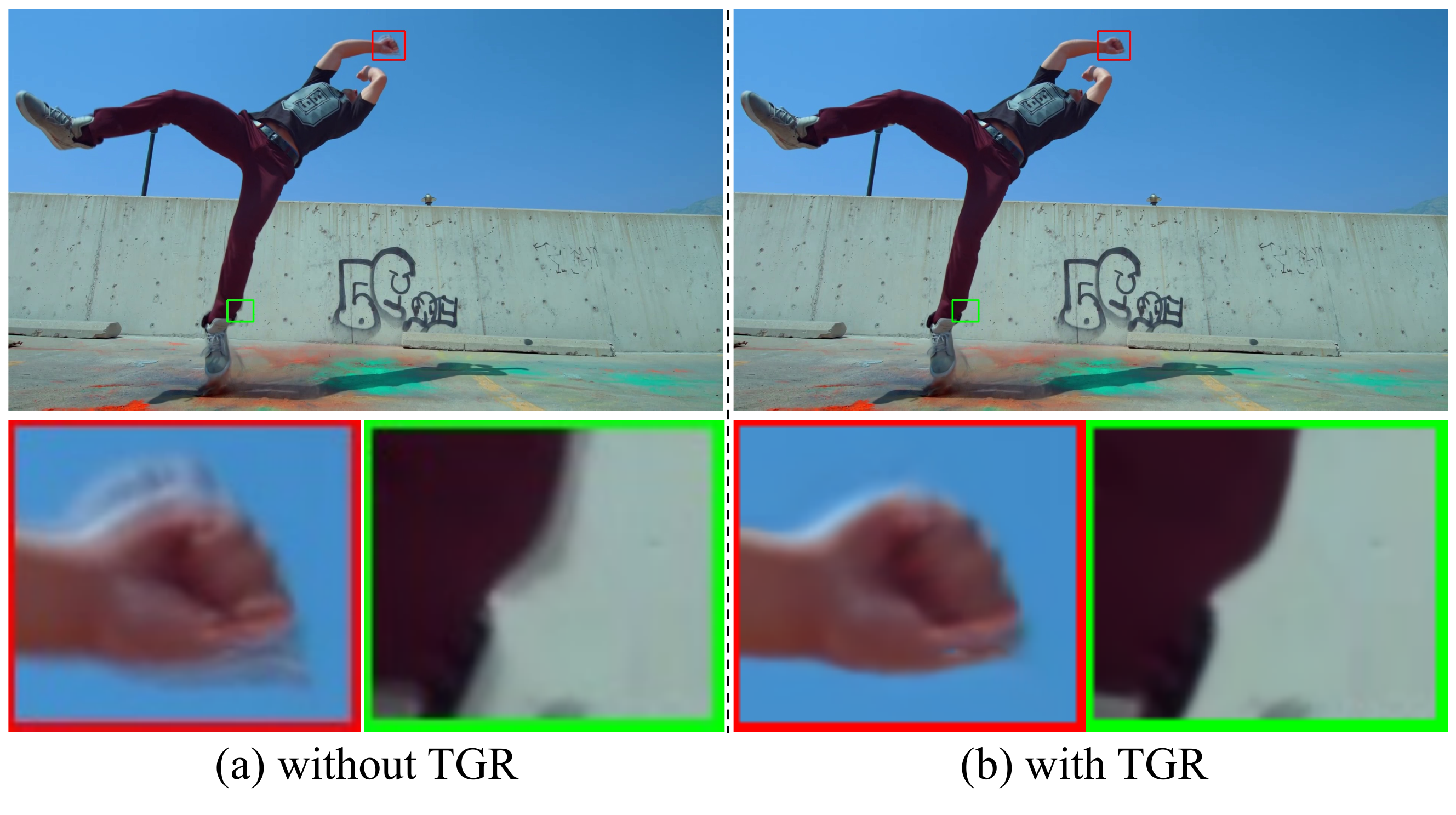}
    \vspace{-0.3in}
    \caption{Ablation Study on Temporal Gated Refinement.}
    \label{fig:ablation_TGR}
    \vspace{-0.2in}
\end{figure}
\vspace{2mm}
\noindent\textbf{Effect of Temporal Gated Refinement}
{ Temporal Gated Refinement (TGR) plays a role of fusion deformable results and degrading artifacts by estimating biases and masks towards HVITB. Figure~\ref{fig:ablation_TGR} shows an example of using TGR. By leveraging TGR, H-VFI alleviates misaligned and blurry results which may produce by simply summing the deformable outputs.}

\vspace{2mm}
\noindent\textbf{Limitation.}
{ We have also noticed that, when the motion range is too large, i.e. larger than the size of deformable kernels at lowest resolution, our method would fail. We believe this problem can be alleviated by increasing the number of hierarchical level in our framework with the drawback of increasing computations.}


%

\section{Conclusion}
In this paper, we propose a simple and effective method, H-VFI, to address the problem of frame interpolation in the presence of large motions. Our H-VFI contributes a hierarchical video interpolation transformer (HVIT) to successively estimate the deformable kernel with learnt residual for predicting the final interpolated frame. Extensive quantitative results and impressive qualitative comparisons clearly show that our proposed method produces competitive and new state-of-the-art frame interpolation results especially in handling complex videos with large motion. 
Furthermore, we contribute a large-scale video dataset at high resolution and high frame rate, with a great variety of scenes and modalities for future research on video frame interpolation and related tasks.

{\small
\bibliographystyle{ieee_fullname}
\bibliography{egbib}

\begin{thebibliography}{10}\itemsep=-1pt

\bibitem{dali}
The nvidia data loading library ({DALI}).
\newblock \url{https://docs.nvidia.com/deeplearning/dali/user-guide/docs/}.

\bibitem{dain2019}
Wenbo Bao, Wei{-}Sheng Lai, Chao Ma, Xiaoyun Zhang, Zhiyong Gao, and
  Ming{-}Hsuan Yang.
\newblock Depth-aware video frame interpolation.
\newblock In {\em {CVPR}}, 2019.

\bibitem{memc2021}
Wenbo Bao, Wei{-}Sheng Lai, Xiaoyun Zhang, Zhiyong Gao, and Ming{-}Hsuan Yang.
\newblock Memc-net: Motion estimation and motion compensation driven neural
  network for video interpolation and enhancement.
\newblock {\em {IEEE} Trans. Pattern Anal. Mach. Intell.}, 43(3):933--948,
  2021.

\bibitem{fruc:tip2018}
Wenbo Bao, Xiaoyun Zhang, Li Chen, Lianghui Ding, and Zhiyong Gao.
\newblock High-order model and dynamic filtering for frame rate up-conversion.
\newblock {\em {IEEE} Trans. Image Process.}, 27(8):3813--3826, 2018.

\bibitem{fruc:trans1996}
Roberto Castagno, Petri Haavisto, and Giovanni Ramponi.
\newblock A method for motion adaptive frame rate up-conversion.
\newblock {\em {IEEE} Trans. Circuits Syst. Video Technol.}, 6(5):436--446,
  1996.

\bibitem{chen2022activating}
Xiangyu Chen, Xintao Wang, Jiantao Zhou, and Chao Dong.
\newblock Activating more pixels in image super-resolution transformer.
\newblock {\em arXiv preprint arXiv:2205.04437}, 2022.

\bibitem{pdwn}
Zhiqi Chen, Ran Wang, Haojie Liu, and Yao Wang.
\newblock Pdwn: Pyramid deformable warping network for video interpolation.
\newblock {\em IEEE Open Journal of Signal Processing}, 2, 2021.

\bibitem{dsepconv2020}
Xianhang Cheng and Zhenzhong Chen.
\newblock Video frame interpolation via deformable separable convolution.
\newblock In {\em {AAAI}}, 2020.

\bibitem{edsc2020}
Xianhang Cheng and Zhenzhong Chen.
\newblock Multiple video frame interpolation via enhanced deformable separable
  convolution.
\newblock {\em {IEEE} Trans. Pattern Anal. Mach. Intell.}, 2021.

\bibitem{choi2020}
Hyomin Choi and Ivan~V. Bajic.
\newblock Deep frame prediction for video coding.
\newblock {\em {IEEE} Trans. Circuits Syst. Video Technol.}, 30(7):1843--1855,
  2020.

\bibitem{cain}
Myungsub Choi, Heewon Kim, Bohyung Han, Ning Xu, and Kyoung~Mu Lee.
\newblock Channel attention is all you need for video frame interpolation.
\newblock In {\em {AAAI}}, 2020.

\bibitem{cdfi2021}
Tianyu Ding, Luming Liang, Zhihui Zhu, and Ilya Zharkov.
\newblock {CDFI:} compression-driven network design for frame interpolation.
\newblock In {\em {CVPR}}, 2021.

\bibitem{superslomo2018}
Huaizu Jiang, Deqing Sun, Varun Jampani, Ming{-}Hsuan Yang, Erik~G.
  Learned{-}Miller, and Jan Kautz.
\newblock Super slomo: High quality estimation of multiple intermediate frames
  for video interpolation.
\newblock In {\em {CVPR}}, 2018.

\bibitem{flavr}
Tarun Kalluri, Deepak Pathak, Manmohan Chandraker, and Du Tran.
\newblock Flavr: Flow-agnostic video representations for fast frame
  interpolation.
\newblock {\em arxiv}, 2021.

\bibitem{adam}
Diederik~P. Kingma and Jimmy Ba.
\newblock Adam: {A} method for stochastic optimization.
\newblock In {\em {ICLR}}, 2015.

\bibitem{kong2022ifrnet}
Lingtong Kong, Boyuan Jiang, Donghao Luo, Wenqing Chu, Xiaoming Huang, Ying
  Tai, Chengjie Wang, and Jie Yang.
\newblock Ifrnet: Intermediate feature refine network for efficient frame
  interpolation.
\newblock In {\em Proceedings of the IEEE/CVF Conference on Computer Vision and
  Pattern Recognition}, pages 1969--1978, 2022.

\bibitem{adacof2020}
Hyeongmin Lee, Taeoh Kim, Tae{-}Young Chung, Daehyun Pak, Yuseok Ban, and
  Sangyoun Lee.
\newblock Adacof: Adaptive collaboration of flows for video frame
  interpolation.
\newblock In {\em {CVPR}}, 2020.

\bibitem{liang2021swinir}
Jingyun Liang, Jiezhang Cao, Guolei Sun, Kai Zhang, Luc Van~Gool, and Radu
  Timofte.
\newblock Swinir: Image restoration using swin transformer.
\newblock In {\em Proceedings of the IEEE/CVF International Conference on
  Computer Vision}, pages 1833--1844, 2021.

\bibitem{cycle2019}
Yu{-}Lun Liu, Yi{-}Tung Liao, Yen{-}Yu Lin, and Yung{-}Yu Chuang.
\newblock Deep video frame interpolation using cyclic frame generation.
\newblock In {\em {AAAI}}, 2021.

\bibitem{eqvi2020}
Yihao Liu, Liangbin Xie, Siyao Li, Wenxiu Sun, Yu Qiao, and Chao Dong.
\newblock Enhanced quadratic video interpolation.
\newblock In {\em {ECCVW}}, 2020.

\bibitem{liu2021Swin}
Ze Liu, Yutong Lin, Yue Cao, Han Hu, Yixuan Wei, Zheng Zhang, Stephen Lin, and
  Baining Guo.
\newblock Swin transformer: Hierarchical vision transformer using shifted
  windows.
\newblock In {\em Proceedings of the IEEE/CVF International Conference on
  Computer Vision (ICCV)}, 2021.

\bibitem{voxelflow2017}
Ziwei Liu, Raymond~A. Yeh, Xiaoou Tang, Yiming Liu, and Aseem Agarwala.
\newblock Video frame synthesis using deep voxel flow.
\newblock In {\em {ICCV}}, 2017.

\bibitem{lu2022video}
Liying Lu, Ruizheng Wu, Huaijia Lin, Jiangbo Lu, and Jiaya Jia.
\newblock Video frame interpolation with transformer.
\newblock In {\em Proceedings of the IEEE/CVF Conference on Computer Vision and
  Pattern Recognition}, pages 3532--3542, 2022.

\bibitem{lucas-Kanade}
Bruce~D. Lucas and Takeo Kanade.
\newblock An iterative image registration technique with an application to
  stereo vision.
\newblock In {\em {IJCAI}}, 1981.

\bibitem{meister2018unflow}
Simon Meister, Junhwa Hur, and Stefan Roth.
\newblock Unflow: Unsupervised learning of optical flow with a bidirectional
  census loss.
\newblock In {\em Proceedings of the AAAI conference on artificial
  intelligence}, volume~32, 2018.

\bibitem{contextaware2018}
Simon Niklaus and Feng Liu.
\newblock Context-aware synthesis for video frame interpolation.
\newblock In {\em {CVPR}}, 2018.

\bibitem{softmaxsplat2020}
Simon Niklaus and Feng Liu.
\newblock Softmax splatting for video frame interpolation.
\newblock In {\em {CVPR}}, 2020.

\bibitem{sepconv2017}
Simon Niklaus, Long Mai, and Feng Liu.
\newblock Video frame interpolation via adaptive separable convolution.
\newblock In {\em {ICCV}}, 2017.

\bibitem{adaconv2017}
Simon Niklaus, Long Mai, and Feng Liu.
\newblock Video frame interpolation via adaptive separable convolution.
\newblock In {\em {ICCV}}, 2017.

\bibitem{bmbc}
Junheum Park, Keunsoo Ko, Chul Lee, and Chang{-}Su Kim.
\newblock {BMBC:} bilateral motion estimation with bilateral cost volume for
  video interpolation.
\newblock In {\em {ECCV}}, 2020.

\bibitem{abme2021}
Junheum Park, Chul Lee, and Chang-Su Kim.
\newblock Asymmetric bilateral motion estimation for video frame interpolation.
\newblock In {\em {ICCV}}, 2021.

\bibitem{imnet2019}
Tomer Peleg, Pablo Szekely, Doron Sabo, and Omry Sendik.
\newblock Im-net for high resolution video frame interpolation.
\newblock In {\em {CVPR}}, 2019.

\bibitem{davis}
F. Perazzi, J. Pont-Tuset, B. McWilliams, L. {Van Gool}, M. Gross, and A.
  Sorkine-Hornung.
\newblock A benchmark dataset and evaluation methodology for video object
  segmentation.
\newblock In {\em {CVPR}}.

\bibitem{reda2022film}
Fitsum Reda, Janne Kontkanen, Eric Tabellion, Deqing Sun, Caroline Pantofaru,
  and Brian Curless.
\newblock Film: Frame interpolation for large motion.
\newblock {\em arXiv preprint arXiv:2202.04901}, 2022.

\bibitem{reda2019}
Fitsum~A. Reda, Deqing Sun, Aysegul Dundar, Mohammad Shoeybi, Guilin Liu,
  Kevin~J. Shih, Andrew Tao, Jan Kautz, and Bryan Catanzaro.
\newblock Unsupervised video interpolation using cycle consistency.
\newblock In {\em {ICCV}}, 2019.

\bibitem{shi2022video}
Zhihao Shi, Xiangyu Xu, Xiaohong Liu, Jun Chen, and Ming-Hsuan Yang.
\newblock Video frame interpolation transformer.
\newblock In {\em CVPR}, 2022.

\bibitem{xvfi2021}
Hyeonjun Sim, Jihyong Oh, and Munchurl Kim.
\newblock Xvfi: extreme video frame interpolation.
\newblock In {\em {ICCV}}, 2021.

\bibitem{aim2020}
Sanghyun Son, Jaerin Lee, Seungjun Nah, Radu Timofte, Kyoung~Mu Lee, Yihao Liu,
  Liangbin Xie, Siyao Li, Wenxiu Sun, Yu Qiao, Chao Dong, Woonsung Park,
  Wonyong Seo, Munchurl Kim, Wenhao Zhang, Pablo~Navarrete Michelini, Kazutoshi
  Akita, and Norimichi Ukita.
\newblock {AIM} 2020 challenge on video temporal super-resolution.
\newblock In {\em Computer Vision - {ECCVW}}, 2020.

\bibitem{raft}
Zachary Teed and Jia Deng.
\newblock {RAFT:} recurrent all-pairs field transforms for optical flow.
\newblock In {\em {ECCV}}, 2020.

\bibitem{tdan2020}
Yapeng Tian, Yulun Zhang, Yun Fu, and Chenliang Xu.
\newblock {TDAN:} temporally-deformable alignment network for video
  super-resolution.
\newblock In {\em {CVPR}}, 2020.

\bibitem{edvr2019}
Xintao Wang, Kelvin C.~K. Chan, Ke Yu, Chao Dong, and Chen~Change Loy.
\newblock {EDVR:} video restoration with enhanced deformable convolutional
  networks.
\newblock In {\em {CVPRW}}, 2019.

\bibitem{emmcvpr2011}
Manuel Werlberger, Thomas Pock, Markus Unger, and Horst Bischof.
\newblock Optical flow guided tv-l video interpolation and restoration.
\newblock In {\em {EMMCVPR}}, 2011.

\bibitem{edge-detect2015}
Saining Xie and Zhuowen Tu.
\newblock Holistically-nested edge detection.
\newblock In {\em {ICCV}}, 2015.

\bibitem{qvi2019}
Xiangyu Xu, Li Si{-}Yao, Wenxiu Sun, Qian Yin, and Ming{-}Hsuan Yang.
\newblock Quadratic video interpolation.
\newblock In {\em {NeurIPS}}, 2019.

\bibitem{vimeo}
Tianfan Xue, Baian Chen, Jiajun Wu, Donglai Wei, and William~T. Freeman.
\newblock Video enhancement with task-oriented flow.
\newblock {\em Int. J. Comput. Vis.}, 127(8):1106--1125, 2019.

\bibitem{yudeep}
Zhiyang Yu, Yu Zhang, Xujie Xiang, Dongqing Zou, Xijun Chen, and Jimmy~S Ren.
\newblock Deep bayesian video frame interpolation.

\bibitem{zoom-to-check2019}
Liangzhe Yuan, Yibo Chen, Hantian Liu, Tao Kong, and Jianbo Shi.
\newblock Zoom-in-to-check: Boosting video interpolation via instance-level
  discrimination.
\newblock In {\em {CVPR}}, 2019.

\end{thebibliography}
}
\clearpage
\section{Structure Details}
\subsection{Hierarchical Video Interpolation Transformer (HVIT)}
We streamline the structure of each level of Hierarchical Video Interpolation Transformer (HVIT) module, shown in Figure~\ref{fig:hvfi}. Following SwinIR~\cite{liang2021swinir}, in each level of HVIT, a convolution layer is first implemented for channel projection. After an HVITB residual, we employed a strided convolution for 2$\times$ downsampling to connect with the next level block. The embedding dimension of HVITB increased from 32 to 64 by enlarging H-VFI to H-VFI-Large.

\begin{figure}[hbp]
    \centering
    \includegraphics[width=\linewidth]{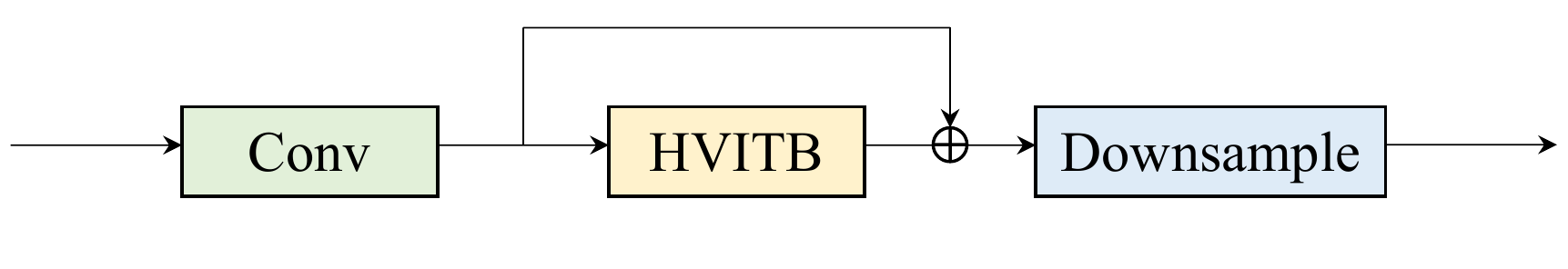}
    \caption{Network structure of each level in HVIT. Downsample refers to convolution with stride of 2.}
    \label{fig:hvfi}
\end{figure}

\subsection{Deformable Head}
\begin{figure*}[htbp]
    \centering
    \includegraphics[width=\linewidth]{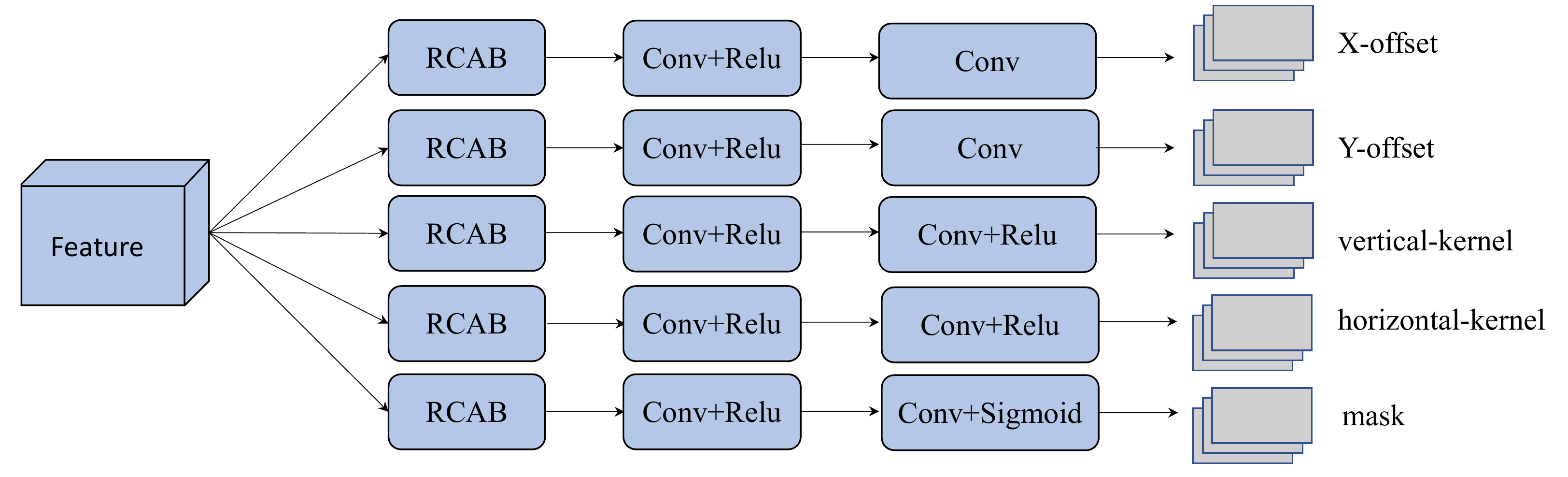}
    \caption{Framework of the proposed Deformable Head}
    \label{fig:dehead}
\end{figure*}

The deformable head shown in Figure~\ref{fig:dehead} is a module consisting of 5 prediction
heads in parallel, which produce the X-offset, Y-offset, vertical kernel, horizontal
kernel and mask respectively for each frame. All branches take the same
feature as input and share the similar structure: residual channel attention blocks
with convolution layers are stacked, followed by another 3×3 convolution layer. For offset
and mask branches, we respectively adopt ReLU and sigmoid as the activation
functions. The output of all branches has the same spatial size with the input
frame, while the number of channels are different. Assuming the deformable ker-
nel size is $n$, The X-offset and the Y-offset are tensors of $n^2$ channels representing
the vertical and horizontal offsets of $n^2$ corresponding pixels, respectively. The
vertical kernel and the horizontal kernel are two tensors of n channels repre-
senting two 1-dimensional (1-D) kernels which are used to approximate one 2-D
kernel as in the DSepConv~\cite{dsepconv2020}. The mask is a tensor of $n^2$ representing the
weights of the corresponding pixels.

\begin{figure}[htbp]
    \centering
    \includegraphics[width=\linewidth]{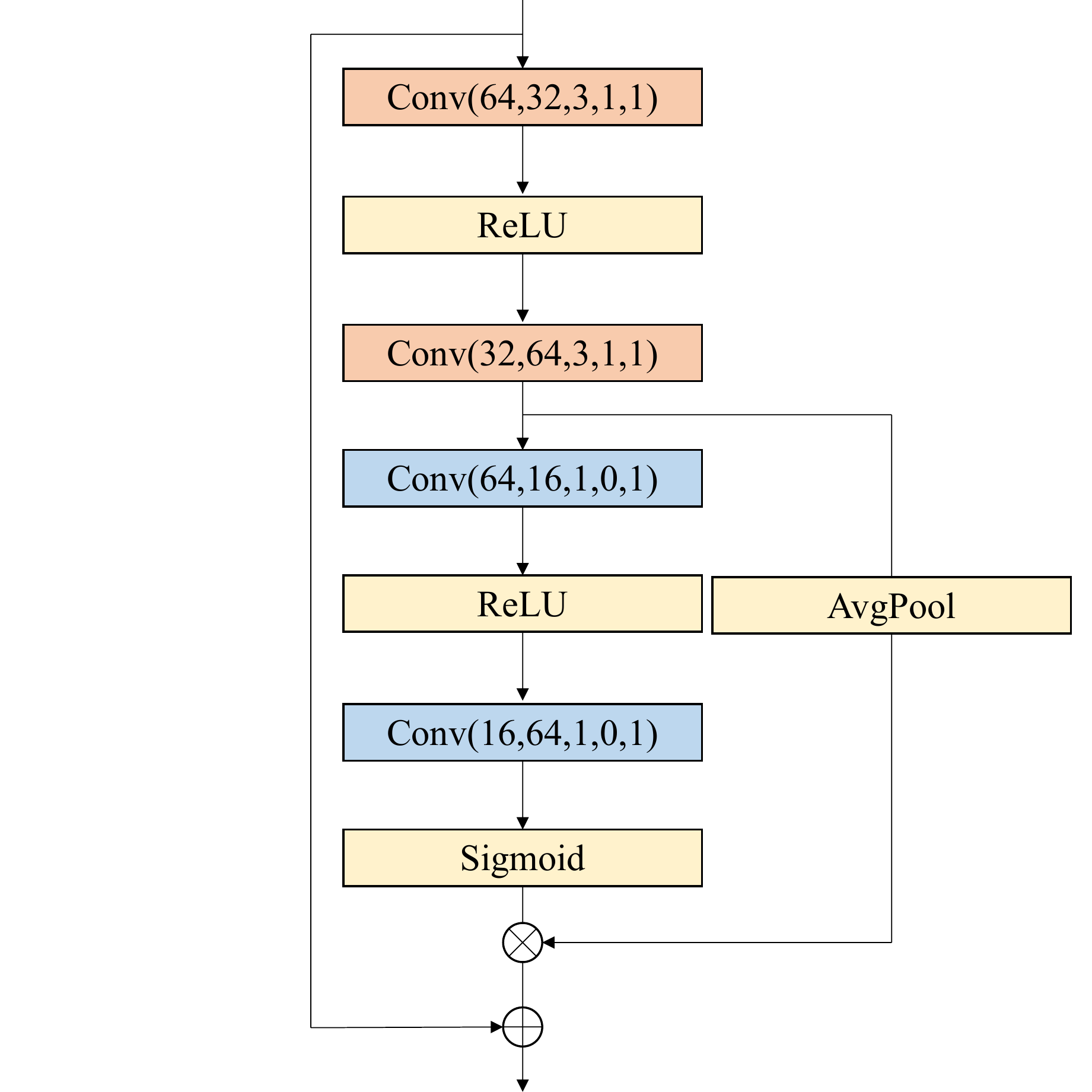}
    \caption{Detail of RCAB, arguments of Conv from left to right are input channels, output channels, kernel size, stride and padding separately.}
    \label{fig:rcab}
\end{figure}

\begin{figure*}[t]
    \centering
    \includegraphics[width=\linewidth]{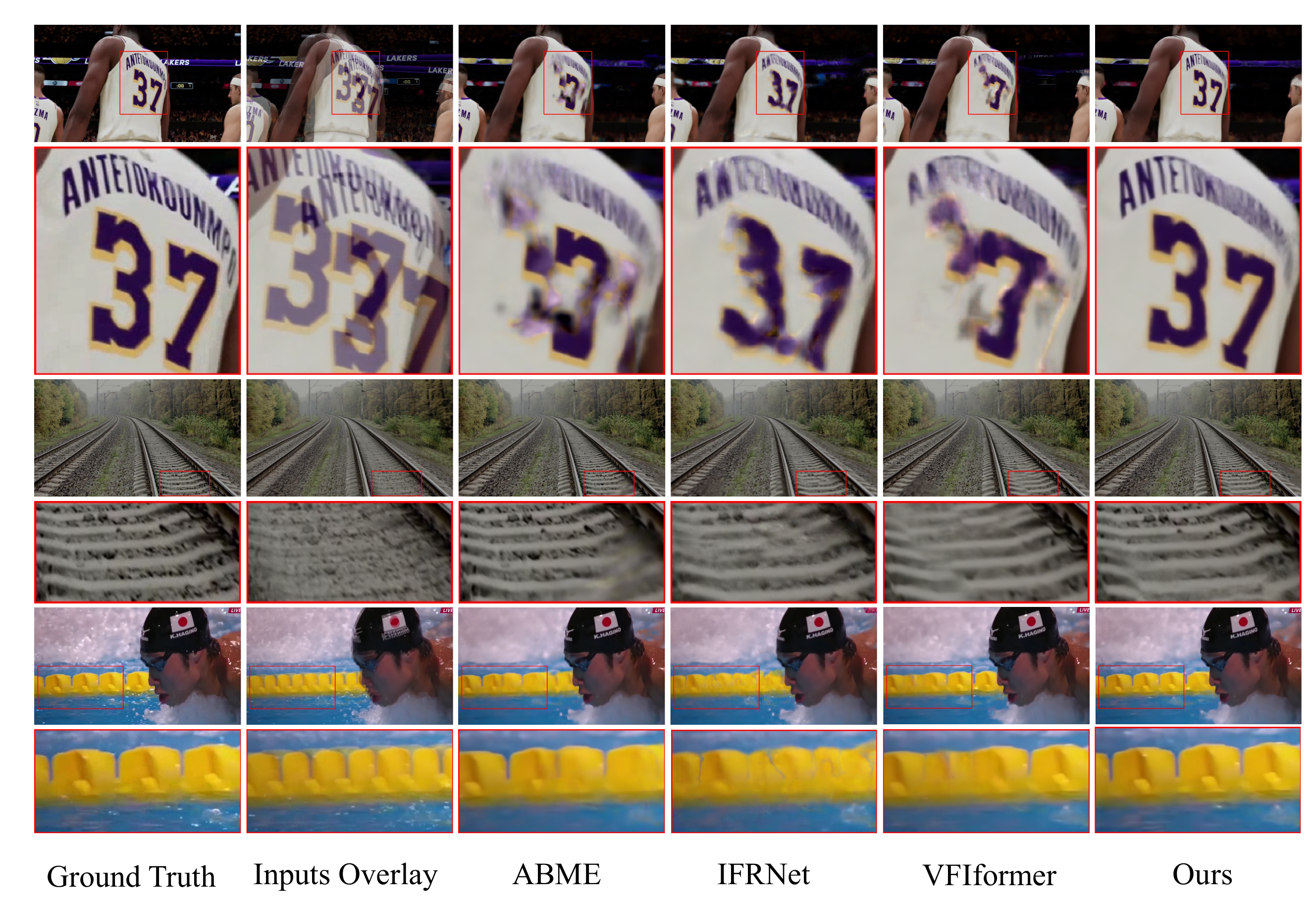}
    \caption{Comparison Results on Youtube200K with ABME, IFRNet, VFIformer and H-VFI}
    \label{fig:more_compare1}
\end{figure*}

\subsection{Residual Channel Attention Block}
We present Residual Channel Attention Block in both HVIT and UDblock. The pipeline detail of one level of RCAB is shown in Figure~\ref{fig:rcab}. For H-VFI and H-VFI-large, each RCAB contains 3 and 4 residual channel attention layers respectively.



\begin{figure*}[htbp]
    \centering
    \vspace{-0.2in}
    \includegraphics[width=\linewidth]{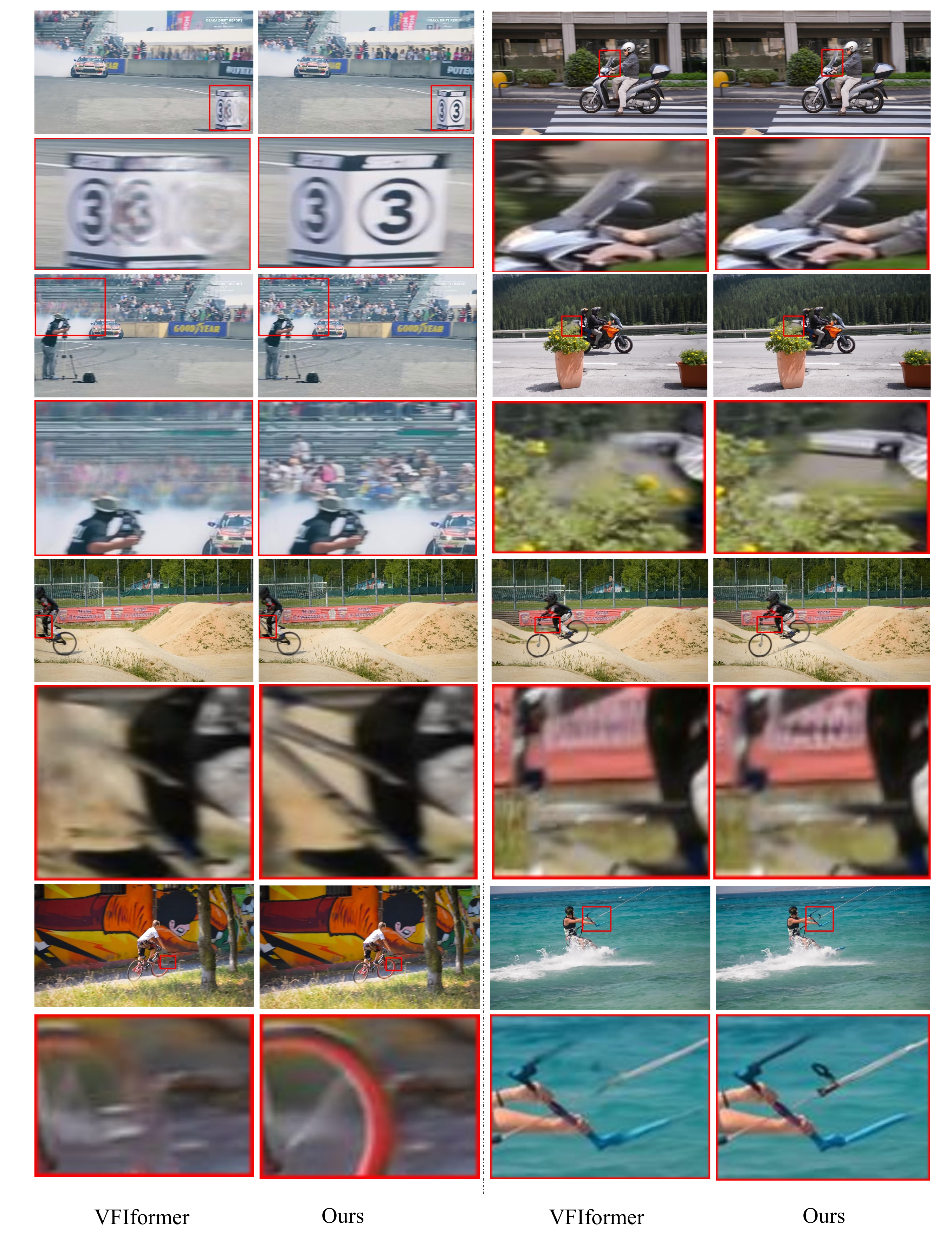}
    \caption{Comparison Results on DAVIS with VFIformer and H-VFI}
    \label{fig:more_compare2}
    \vspace{-0.2in}
\end{figure*}

\section{More Comparsion Results}
We  show further qualitative comparisons with H-VFI, VFIformer~\cite{lu2022video}, ABME~\cite{abme2021} and IFRNet~\cite{kong2022ifrnet} in Figure~\ref{fig:more_compare1} and Figure~\ref{fig:more_compare2}. The  examples are selected from the DAVIS~\cite{davis} dataset and our proposed Youtube200K dataset for fairly comparison. By leveraging our hierarchical architecture and kernel updating strategy, our H-VFI can better handle a large motion of various objects or in the presence of occlusion than existing state-of-the-art methods.
\end{document}